\documentclass[10pt,twocolumn,letterpaper]{article}
\pdfoutput=1

\usepackage{iccv}
\usepackage{times}
\usepackage{epsfig}
\usepackage{graphicx}
\usepackage{amsmath}
\usepackage{amssymb}

\usepackage[flushleft]{threeparttable}
\usepackage{colortbl}
\usepackage{cite}
\usepackage{amsfonts}
\usepackage{amsthm}
\usepackage{algorithm,algpseudocode}
\usepackage{multirow}
\usepackage{enumitem}
\usepackage{textcomp}
\usepackage{subcaption}
\usepackage{xcolor}
\usepackage{bm}

\usepackage[pagebackref=true,breaklinks=true,colorlinks,bookmarks=false]{hyperref}

\iccvfinalcopy 

\def\iccvPaperID{****} 

\ificcvfinal\fi

\algrenewcommand\algorithmicindent{1.0em}
\algnewcommand{\Inputs}[1]{%
  \State \textbf{Inputs:}
  \Statex \hspace*{\algorithmicindent}\parbox[t]{.8\linewidth}{\raggedright #1}
}
\algnewcommand{\Outputs}[1]{%
  \State \textbf{Outputs:}
  \Statex \hspace*{\algorithmicindent}\parbox[t]{.8\linewidth}{\raggedright #1}
}
\algnewcommand{\Initialize}[1]{%
  \State \textbf{Initialize:}
  \Statex \hspace*{\algorithmicindent}\parbox[t]{1.2\linewidth}{\raggedright #1}
}

\newcommand{\bd}{{\bf d}}
\newcommand{\bestd}{{\bf d}^*}
\newcommand{\btheta}{{\boldsymbol{\theta}}}

\newtheorem{theorem}{Theorem}
\newtheorem{coro}[theorem]{Corollary}
\newcommand{\argmin}{\mathop{\rm arg~min}\limits}

\newcommand{\ASE}{\mathrm{ASR}}
\newcommand{\Ours}[0]{TA{\it k}S\xspace}
\definecolor{Gray}{gray}{0.9}
\newcommand{\lowacc}[1]{{\cellcolor{Gray} #1}}

\begin{document}

\title{No Regret Sample Selection with Noisy Labels}

\author{Heon Song\textsuperscript{*,†} \hspace{1cm} Nariaki Mitsuo\textsuperscript{*} \hspace{1cm} Seiichi Uchida\textsuperscript{*} \hspace{1cm} Daiki Suehiro\textsuperscript{*,†}\\
\textsuperscript{*}Kyushu University, Japan \hspace{1cm} \textsuperscript{†}RIKEN AIP, Japan\\
{\tt\small \{heon.song@human., uchida@, suehiro@\}ait.kyushu-u.ac.jp}
}

\maketitle
\ificcvfinal\thispagestyle{empty}\fi

\begin{abstract}
Deep neural networks (DNNs) suffer from noisy-labeled data because of the risk of overfitting. To avoid the risk, in this paper, we propose a novel DNN training method with sample selection based on adaptive $k$-set selection, which selects $k~(< n)$ samples with a small noise-risk from the whole $n$ noisy training samples at each epoch. It has a strong advantage of guaranteeing the performance of the selection theoretically. Roughly speaking, a regret, which is defined by the difference between the actual selection and the best selection, of the proposed method is theoretically bounded, even though the best selection is unknown until the end of all epochs. The experimental results on multiple noisy-labeled datasets demonstrate that our sample selection strategy works effectively in the DNN training; in fact, the proposed method achieved the best or the second-best performance among state-of-the-art methods, while requiring a significantly lower computational cost.
\end{abstract}

\section{Introduction}
Deep neural networks (DNNs) require a large number of ``correctly-labeled'' samples to achieve high-performance classification~\cite{zhang2016understanding}. However, it is practically difficult for many CV/PR-related datasets to guarantee the correctness of the attached labels~\cite{frenay2013classification,sukhbaatar2015training}. For example, datasets annotated via crowd-sourcing~\cite{yan2014learning,yu2018learning} often 
become noisy-labeled samples, which contain a certain number of samples with incorrect labels. Beyer et al.~\cite{beyer2020we} point out that ImageNet contains several incorrectly labeled samples, even though each sample in ImageNet is labeled by a careful majority voting scheme among at least 10 workers~\cite{deng2009imagenet}. Datasets created by automatic data collection and annotation, such as Clothing1M~\cite{xiao2015learning}, will also contain a large number of incorrectly labeled samples.
\par
\begin{figure*}[t]
\begin{center}
\includegraphics[width=1\textwidth]{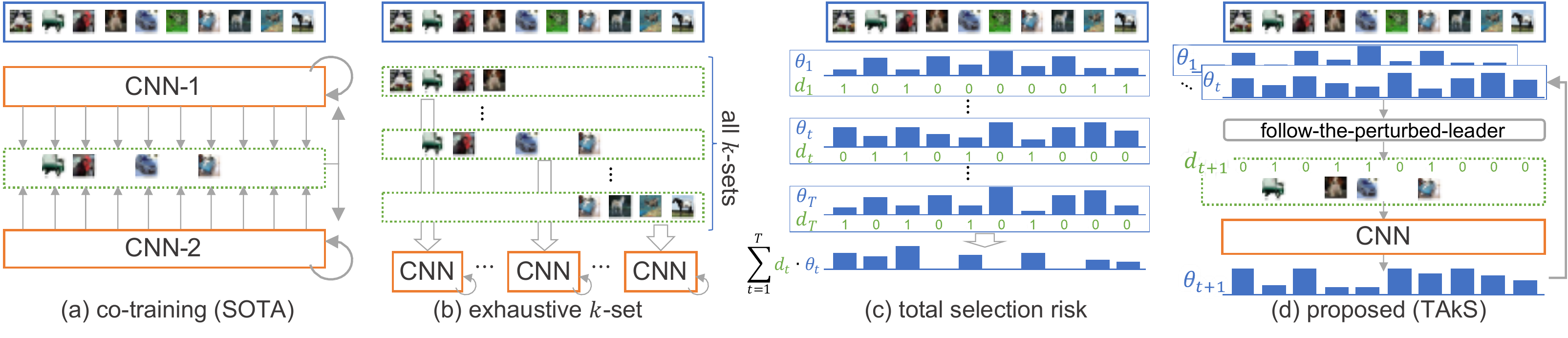}\\[-3mm]
\caption{(a)~Co-training. (b)~The most naive and computationally-intractable $k$-set selection method. 
 (c)~Total selection risk. (d)~The proposed method based on adaptive $k$-set selection. The vectors $\btheta_t$ and $\bd_t$ are the noise-risk vector and the $k$-hot vector at epoch $t$, respectively. In (d), the perturbation mechanism in the $k$-set selection is omitted for simplicity.}
\label{fig:subset-selection}
\end{center}
\vskip -5mm
\end{figure*}

A possible remedy for noisy-labeled samples is {\em sample selection}, which is a method to select a clean subset of training samples  (i.e., correctly labeled training samples) from the whole sample set.
A typical strategy is co-training, which compares the results from two DNNs to detect the incorrectly labeled samples, as shown in Fig.~\ref{fig:subset-selection}~(a).
This strategy has achieved state-of-the-art performance so far but requires extra computations due to its coupled-DNN structure. Another drawback is that it has no theoretical guarantee on its performance. 
\par
We can consider another sample selection strategy by using the idea of {\em $k$-set}, which is a subset of $k$ samples. Fig.~\ref{fig:subset-selection}~(b) shows the most naive $k$-set-based sample selection strategy; after generating all $\binom{n}{k}$ $k$-sets exhaustively from the whole training set with $n$ samples, $\binom{n}{k}$ DNNs are trained independently with individual $k$-sets. If there are more than $k$ clean samples, some DNNs are trained only by clean samples and thus avoid the degradation by incorrectly labeled samples. However, this naive and exhaustive $k$-set strategy is obviously intractable in practical scenarios with larger $n$ and $k$.
\par

This paper proposes a method where $k$-set selection and DNN training by the selected $k$ samples are performed alternatively. Let $\bd_t$ denote the $n$-dimensional $k$-set vector showing the selection at epoch $t$ and $\btheta_{t-1}$ denote the $n$-dimensional {\em noise-risk vector} whose $i$th element is the noise-risk of the $i$th training sample, which is assessed at $t-1$. The noise-risk is the likelihood that the sample is incorrectly labeled. In the ideal case when $\btheta_t$ is highly reliable, we can select the $k$-set $\bd_t$ that minimizes $\bd_t\cdot\btheta_{t-1}$. 
However, this greedy strategy does not work because of the difficulty of having a reliable $\btheta_t$ during the training process.
\par
We, therefore, design our method to suppress the {\em total selection risk}, which is defined as $\sum_{t=1}^T\bd_t\cdot\btheta_t$, as shown in Fig.~\ref{fig:subset-selection}~(c). By achieving a smaller total selection risk at $T$, we can believe that DNN is trained with less incorrectly-labeled samples {\em on average} during the entire training process. Due to the online (i.e., non-predictable) nature of the training process, one may imagine that it is difficult to determine $\bd_t$ at $t$ while guaranteeing a smaller total selection risk. We prove that it is still possible to achieve a total selection risk smaller than a theoretical bound, based on the theory of {\em adaptive $k$-set selection}~\cite{warmuth2008randomized} with {\em Follow-the-Perturbed Leader} (FPL) algorithm, as discussed later. Hereafter we call the proposed method {\em training DNN
with sample selection based on adaptive k-set selection} (\Ours).
\par
Fig.~\ref{fig:subset-selection}~(d) shows \Ours, which has the following three advantages.
The first and most important advantage is its theoretical support. Specifically, its performance is theoretically guaranteed in terms of {\em regret}. Regret is a performance measure and is often used in theoretical machine learning research. In our $k$-set selection task, the regret $R_T$ is defined as the difference between the actual total selection risk and the minimum total selection risk with the best $k$-set $\bestd$. (See Eq.~(\ref{eq:regret}) for the formal definition of $R_T$.) Although $\bestd$ is unknown until $T$, $R_T$ of \Ours is upper-bounded; this means that the $k$-set selection by \Ours is not far from $\bestd$, even under the uncertain online nature of the selection process.
\par 
The second advantage is its computational efficiency. \Ours selects an appropriate $k$-set $\bd_t$ from all the $\binom{n}{k}$ $k$-set candidates with just $O(n)$ time complexity at each $t$. In addition, \Ours trains a single DNN, whereas the state-of-the-art sample selection methods (i.e., the co-training methods) need to train coupled-DNNs as noted above. 
\par
Finally, \Ours achieves the best or near-best classification accuracy, as shown by our experimental results on several benchmark datasets, such as  MNIST~\cite{lecun2010mnist}, CIFAR-10, CIFAR-100~\cite{krizhevsky2009learning}, and Clothing1M~\cite{xiao2015learning}, which have often been used in the past studies of noisy-labeled samples. Under various noise settings, \Ours achieves the best or the second-best accuracy and time efficiency among various state-of-the-art sample selection methods \cite{han2018co,yu2019does,wei2020combating}.
\par
The summary of our main contributions are as follows:
\begin{itemize}
    \item For learning with noisy-labels, we propose a novel and efficient method, \Ours, which tries to suppress the total selection risk based on the theory of adaptive $k$-set selection. To the best of our knowledge, \Ours is the first method that utilizes the idea of adaptive $k$-set selection for noisy-labeled samples. 
    \item We discuss how the theoretical support of \Ours is useful for selecting a promising $k$-set from noisy-labeled samples. We also prove how the performance of \Ours is guaranteed in terms of the total selection risk. 
    \item The experimental results on multiple noisy-labeled datasets show that \Ours achieved the best or the second-best performance in not only classification accuracy but also time efficiency among state-of-the-art methods. We also confirmed that \Ours selects more clean samples than the other comparative methods. 
\end{itemize}

\section{Related Work\label{subsec:related}}
\subsection{Learning with noisy labels}
\paragraph{Sample selection strategies for noisy-labeled samples}
Among the various methods for dealing with noisy-labeled samples, one of the most popular choices is sample selection, which selects clean sample candidates from the whole training sample set. The clear merit of sample selection is that it can totally exclude the unexpected effect of the non-selected incorrectly labeled samples. Moreover, 
it is also useful to remove irrelevant samples before reusing the sample set for another task.
\par
A widely used sample selection strategy for DNN training is to select samples with small training loss (such as cross-entropy loss for the classification task). ITML~\cite{shen2019learning} considers the samples with a small training loss in the initial epochs to be clean samples. Therefore, it selects samples with small training loss at the end of every epoch and trains the DNN with the samples.
 Unlike this strategy, iterative learning~\cite{wang2018iterative} considers incorrectly labeled samples as outliers and removes the noise using the outlier detection algorithm~\cite{breunig2000lof}. 
\par
Recent state-of-the-art methods take the strategy of co-training, which trains coupled-DNNs simultaneously for clean sample selection~\cite{malach2017decoupling,jiang2018mentornet,han2018co,yu2019does,wei2020combating}. Decoupling~\cite{malach2017decoupling} selects the samples that the two classifiers predict differently and uses them for training the classifiers. In Co-teaching~\cite{han2018co}, each of the two classifiers selects the samples with small training loss and then exchanges the samples for training the other classifier. Co-teaching+~\cite{yu2019does} uses both ideas of Decouple and Co-teaching. JoCoR~\cite{wei2020combating} tries to make the classification results of two classifiers closer to each other while selecting samples with a small loss. Although these state-of-the-art methods have shown outstanding performance, their coupled-DNN structure requires almost double the computation time and resources than a single DNN training. Moreover, their theoretical justification has not yet been provided.
%
\paragraph{Other strategies for noisy-labeled data}
One of the strategies to train a DNN with noisy-labeled data is estimating a label transition matrix~\cite{natarajan2013learning,menon2015learning,patrini2017making,xia2019anchor}. The label transition matrix is a matrix that represents the probability of a label flipping from one class to another. F-correction~\cite{patrini2017making} is a popular method that estimates the label transition matrix to improve the robustness of DNN. It estimates the matrix using a well-trained DNN with noisy-labeled data and trains another DNN by weighting losses using the estimated matrix.
This strategy is theoretically well-motivated. However, it is practically difficult to estimate the label transition matrix with many classes. Consequently, the practical performance is not so high, compared with the above sample selection-based methods. \par
Another strategy is to utilize additional ``clean'' data~\cite{li2017learning,veit2017learning,ren2018learning, rodrigues2018deep,yao2020searching}. Veit et al.~\cite{veit2017learning} used clean data to train a label cleaning network. Using the label cleaning network, it is possible to clean up noisy-labeled data and train a DNN with the cleaned data. Ren et al.~\cite{ren2018learning} proposed a weighting algorithm to learn with noisy-labeled data. It weighs the training data to reduce the loss of clean validation data. Rodrigues et al.~\cite{rodrigues2018deep} proposed a crowd layer that contains annotator-specific functions. By assigning reliability to the functions, unreliable annotators and annotators' bias can be avoided. Yao et al.~\cite{yao2020searching} designed a method for improving the existing sample selection method using the AutoML technique. Using clean validation data and its loss, the method searches a proper selection schedule automatically. However, these methods require a lot of time and resources to collect clean data or labels from multiple annotators.

\subsection{Adaptive $k$-set selection}
\label{subsec:kset}
\Ours utilizes the idea of {\em adaptive $k$-set selection}~\cite{warmuth2008randomized}, which is a task of selecting some good $k$ elements from $n$ elements {\em repeatedly}. Roughly speaking, the task is to select a good $k$-set of elements adaptively under the assumption that the temporarily optimal selection can dynamically change in the repeating.
\par
The adaptive $k$-set selection mainly focuses on theoretical research, and its applicability to real-world tasks has not been demonstrated.
In fact, it was originally proposed for theoretical research for the online PCA task~\cite{warmuth2008randomized}. Then, many extended research studies have been proposed~\cite{koolen2010hedging,suehiro2012online,cohen2015following}, and it is still the subject of theoretical research. Therefore, this paper is the first attempt at using adaptive $k$-set selection for noisy-labeled data.\par

\section{Training DNN with Sample Selection Based on Adaptive $k$-set Selection (\Ours)}
\begin{algorithm}[t]
\caption{\Ours}
\label{alg:main}
\begin{algorithmic}[1]
\Inputs{Training samples $(x_1, y_1), \ldots, (x_n, y_n)$, total epochs $T$, initial DNN $f_0$, $k$, $\eta > 0$}
\Outputs{$f_{T}$: trained DNN}
\Initialize{Set $\bd_{1}$ as an $n$-dimensional $k$-hot vector}
\For {epochs $t=1, \ldots, T$}
 \State Obtain $f_t$ by training $f_{t-1}$ with the $k$-set, $\bd_{t}$.
  \State Update the noise-risk vector
  $\btheta_t = (\theta_{1,t}, \ldots, \theta_{n,t}$).
  \State Sample the perturbation ${r}_{i} \sim \mathcal{N}(0,1)$ for all $i$.
  \State Select the next $k$-set of samples: \begin{align}\label{align:FPLupdate}
  \bd_{t+1}=\argmin_{\bd \in\mathcal{D}} \left(\sum^{t}_{\tau=1} \bd \cdot \btheta_\tau + \eta \bd \cdot \boldsymbol{r}  \right)
  \end{align}
\EndFor
\end{algorithmic}
\end{algorithm}
\subsection{Basic settings}
We assume the task of training a DNN-based classifier with noisy-labeled samples. Let $x_1,\ldots, x_n$  be training samples and $y_1, \dots, y_n$ their labels. Some samples are labeled incorrectly.
The goal is to obtain a DNN-based classifier $f_T$ after $T$ epochs and achieve high classification accuracy on a test sample set. Note that, in the popular setup of noisy-labeled samples, the test sample set is not noisy to facilitate the final classification evaluation easier. We also follow the same setup.
\par
Letting $f_t$ denote the DNN after epoch $t$ and $\mathcal{D}$ denote the set of all possible $k$-sets, the basic procedure of \Ours at epoch $t$ is described as the following three steps (recall Fig.~\ref{fig:subset-selection}~(d)):
\begin{enumerate}
\item Select a good $\bd_{t}$ from $\mathcal{D}$ by a selection algorithm.
\item Train $f_{t-1}$ with the $k$ samples in $\bd_t$ and get $f_t$.
\item Assess the noise-risk vector $\btheta_t$.
\end{enumerate}
The details of \Ours based on the procedure are shown in Algorithm~\ref{alg:main} and described in the later sections. Specifically, for Step 1, we will show a theoretical definition of what is a good selection in Sec.~\ref{subsec:total_risk}. We then show a selection algorithm in Sec~\ref{subsec:algo}.
For Step 3, we can use an arbitrary approach, although we expect that 
the $i$th element $\theta_{i,t}$ of the noise-risk vector $\btheta_t \in [0,1]^n$ becomes larger when the noise-risk of the $i$th sample is higher at epoch $t$. Sec.~\ref{subsec:noise-risk-vector} gives a reasonable way to assess $\btheta_t$. Note that $\btheta_t$ is assessed after training the DNN at Step 2; in other words, the previous noise-risk vectors $\btheta_{1}, \ldots,\btheta_{t-1}$ can be used for selecting $\bd_{t}$. 
\par
\subsection{Total selection risk and adaptive $k$-set selection}
\label{subsec:total_risk}
To achieve high classification performance by the DNN, it is better to select $\bd_{t}$ with less incorrectly-labeled samples at each epoch $t$. 
The greedy selection of $\bd_t$ with minimum noise-risk (i.e., $\min_{\bd_t \in \mathcal{D}} \bd_t \cdot \btheta_{t-1}$) at each epoch $t$ is not reasonable; this is because the noise-risk vector $\btheta_{t-1}$ might be quite unstable during training the DNN. For example, $\theta_{i,t}$ can be very large even if $\theta_{i,t-1}$ is very small. It might happen in the DNN training where just one epoch training makes a big change in the evaluation of the $i$th sample.\par
To deal with this instability, \Ours aims to suppress the total selection risk $\sum_{t=1}^T \bd_t \cdot \btheta_t$; this means \Ours aims to suppress the {\em average} selection risk during $t=1,\ldots, T$. This aim seems reasonable but not straightforward because it is non-predictable at each $t < T$ that the selection at $t$ is surely good for a smaller total selection risk, due to the high instability of $\btheta_t$. In other words, we cannot have any ``assumption'' for predicting the total selection risk from the current selection at $t$.
\par
Therefore, we consider the selection task as \emph{adaptive $k$-set selection}~\cite{warmuth2008randomized}.
Adaptive $k$-set selection is defined as a task in which the $k$-set selection and its evaluation are alternated repeatedly, and thus similar to our selection procedure.
The theoretical goal of the adaptive $k$-set selection is to select $\bd_1, \ldots, \bd_T$ with a guarantee of the small total selection risk for {\em any} $\btheta_1,\ldots, \btheta_T$. 
In other words, the adaptive $k$-set selection task assumes even a very tough situation where $\btheta_t$ can change drastically at every $t$ and thus 
is not predictable.  
\par

An important and general theoretical highlight of the adaptive $k$-set selection task is that its \emph{regret} $R_T$ can be upper bounded as:
\begin{align}
\label{eq:regret}
R_T = \sum_{t=1}^T \bd_t \cdot \btheta_t
- \min_{\bd \in \mathcal{D}} \sum_{t=1}^T
\bd \cdot \btheta_t, 
\end{align}
for {\em any} $\btheta_t$ (i.e., for any tough situation). 
The first term is the total selection risk.
The second term is the total selection risk when we use {\em the best selection} $\bestd = \mathrm{argmin}_{\bd} \sum_{t=1}^T
\bd \cdot \btheta_t$ at all $T$ rounds. That is, regret means the relative performance (i.e., total selection risk) difference between the best selection and actual selection.
The upper-bound of $R_T$ depends on the algorithm to select $\bd_t$ at each round. Thus, if we use a selection algorithm that gives a smaller upper bound, we can expect a smaller $R_T$, that is, good performance similar to the best selection $\bestd$. 
\par
The following three points should be emphasized regarding the above regret upper bound. First, $R_T$ can be bounded theoretically even though the best selection $\bestd$ is only known after the end of the $T$ rounds and thus unknown when we decide $\bd_t$ at each $t<T$. Second, $R_T$ is still bounded even though the noise-risk vector $\btheta_t$ fluctuates drastically along with $t$.
Third, the upper bound of $R_T$ depends on the choice of the selection algorithm, as noted above.\par
It should be emphasized again that adaptive $k$-set selection is a general online selection task, and these theoretical supports hold in arbitrary $k$-set selection tasks. Therefore these supports also hold our sample selection task for training a DNN. Especially, having a lower regret bound is very meaningful for our task because it indicates that we can expect a performance similar to the best selection $\bestd$ in terms of the total selection risk. (This point is further supported by Corollary 2.)



\subsection{Adaptive $k$-set selection using FPL}
\label{subsec:algo}
The remaining issue is the choice of the sample selection algorithm that gives a lower regret bound. For \Ours, we employ the {\em Follow-the-Perturbed-Leader} (FPL)~\cite{cohen2015following} as a selection algorithm for adaptive $k$-set selection, because it gives a good regret bound~\cite{warmuth2008randomized,koolen2010hedging,suehiro2012online,cohen2015following}.
FPL is an improved version of the {\em Follow the Leader} (FTL) algorithm~\cite{hazan2016introduction}, which selects the ``current best'' $k$-set $\bestd_t$ at epoch $t$ (i.e., $\bestd_t=\mathrm{argmin}_{\bd} \sum_{\tau=1}^t
\bd \cdot \btheta_\tau$.) Although the idea of FTL seems reasonable, it has already been proved that FTL can overfit to tough $\btheta_1,\ldots, \btheta_T$ and cannot have a meaningful regret bound~\cite{hazan2016introduction}. In contrast, 
FPL selects $k$ samples based on the perturbed selection risk, in order to avoid FTL's overfitting and have the regret bound. More precisely,
Steps 7 and 8 in Algorithm~\ref{alg:main} show the $k$-set selection algorithm based on FPL, where $\boldsymbol{r} = (r_1, \ldots, r_n)$ is a random vector for the perturbation~\cite{kalai2002geometric,cohen2015following}.
\par
\subsection{Efficiency and versatility}

The major computational cost of \Ours comes from the sample selection steps (Steps 6-8 in Algorithm~\ref{alg:main}) and the DNN training cost with $k$ samples (Step 5). 
The former is dominated by the minimization operation in Eq.~(\ref{align:FPLupdate}) and efficiently performed just with $O(n)$ computations despite $|\mathcal{D}|=\binom{n}{k}$.
The term to be minimized is rewritten as follows:
$\left(\sum^{t}_{\tau=1} \bd \cdot \btheta_\tau + \eta\bd \cdot \boldsymbol{r}  \right)=\left(\sum^{t}_{\tau=1} \btheta_\tau + \eta\boldsymbol{r}\right)\cdot\bd$.
Therefore, finding the optimal $\bd$ in Eq.~(\ref{align:FPLupdate}) is equivalent to finding the elements with the $k$ least perturbed total selection risk in $\left(\sum^{t}_{\tau=1} \btheta_\tau + \eta\boldsymbol{r}\right)$ and requires $O(n)$ computations for finding the top-$k$ elements from $n$ elements~\cite{musser1997introspective}. Consequently, \Ours works efficiently with just $O(n)$ computations and thus can deal with large training sample sets.  
\par
Consequently, the latter (Step 5) requires more computations than the former (Steps 6-8) in practice; however, even the latter computation is far more efficient than the standard DNN training. This is because \Ours requires the DNN training cost with $k~(<n)$ samples instead of $n$ samples. 
\par
Algorithm~\ref{alg:main} also shows the versatility of \Ours; we can employ an arbitrary DNN. In the later experiment, we use several different neural networks in \Ours. 


\section{Theoretical Support of \Ours\label{sec:theo}}
In this section, we show that \Ours is theoretically supported by regret-based analysis.
Note that our sample selection method generally works
with the theoretical support for any sequences of the noise-risk vectors.
In other words, to receive the benefit of the theoretical support, we do not need to care how the noise-risk vector is calculated. 
This is because the regret is defined as a comparative difference between the total risk of actual selections and the total risk of the best selections (see Eq.~(\ref{eq:regret})).
The practical selection error (i.e., how much incorrectly labeled data we select) only depends on the absolute selection risk of the best $k$-set selection, which is decreased when we use some good noise-risk calculation (see Sec.~\ref{subsec:noise-risk-vector}).
\subsection{Regret bound of \Ours}
The regret bound of \Ours is directly derived from the following  general theorem~\cite{cohen2015following}:
\begin{theorem}[\cite{cohen2015following}]
\label{theo:rb}
Let $k \in \{1, \ldots, n-1\}$.
By setting $\eta = \sqrt{kT}$~\footnote{\label{footnote:eta}The parameter $\eta$ is theoretically set as $\sqrt{kT}$, for guaranteeing the worst-case performance by using $kT$, i.e., the worst total selection risk. In practice, we can achieve much smaller regret by setting a smaller $\eta$. In our experiment, we set $\eta$ like $\eta=10^{-3}\sqrt{kT}$.},
the regret of the adaptive $k$-set selection algorithm with FPL
is upper bounded as follows:
\begin{align}
  \mathbb{E}[R_T] \leq 2\sqrt{2kT \ln \binom{n}{k}},
\end{align}
where the expectation comes from the randomness of FPL.
\end{theorem}
This theorem says that the regret, which is the performance difference between the selected $k$-sets and the best $k$-set, is bounded by a constant value given by $T$, $n$, and $k$, 
and the bound is not exploded with the huge $\binom{n}{k}$ choices of $k$-sets but just relative to $\sqrt{\ln\binom{n}{k}}$.


\addtolength{\tabcolsep}{-2pt}
\begin{table*}[t]
\centering
\begin{threeparttable}
\caption{Average test accuracy (\%) over the last 10 epochs and training time .\vspace{-2mm}}
\label{table:accuracy}
\begin{small}
\begin{tabular}{c|c|c|c|c|c|c|c|c|c|c|c|c}
\hline
\multirow{2}{*}{(Dataset) Noise} & Standard & F-correction & \multicolumn{2}{c|}{Decouple} & \multicolumn{2}{c|}{Co-teaching} & \multicolumn{2}{c|}{Co-teaching+} & \multicolumn{2}{c|}{JoCoR}
& \multicolumn{2}{c}{\Ours(Ours)} \\
  & Acc & Acc & Acc & $\times$T & Acc & $\times$T  & Acc & $\times$T  & Acc & $\times$T   & Acc & $\times$T \\
\hline\hline
(MNIST) Symmetric-20\% & 78.67 & 90.71 & 94.74 & \textbf{1.10} & 94.52 & 1.60 & 97.77 & 1.61 & {\color{blue} \textit{97.88}} & 1.47 & {\color{red} \textbf{97.94}} & 1.19 \\
(MNIST) Symmetric-50\% & 51.22 & 77.36 & 66.77 & 1.46 & 89.50 & 1.59 & 95.67 & 1.67 & {\color{blue} \textit{95.90}} & 1.49 & {\color{red} \textbf{97.17}} & \textbf{0.97} \\
(MNIST) Symmetric-80\% & 22.43 & 51.16 & 27.42 & 1.50 & 78.52 & 1.59 & 66.13 & 1.73 & {\color{blue} \textit{88.53}} & 1.49 & {\color{red} \textbf{92.32}} & \textbf{0.81} \\
(MNIST) Asymmetric-40\% & 78.97 & 88.99 & 82.04 & 1.35 & 90.21 & 1.59 & 92.48 & 1.65 & {\color{blue} \textit{93.91}} & 1.50 & {\color{red} \textbf{95.77}} & \textbf{1.22} \\
\hline
(CIFAR-10) Symmetric-20\% & 68.92 & 74.21 & 69.95 & 1.61 & 78.16 & 1.98 & 78.68 & 2.00 & {\color{red} \textbf{85.75}} & 1.73 & {\color{blue} \textit{83.90}} & \textbf{0.99} \\
(CIFAR-10) Symmetric-50\% & 41.93 & 52.68 & 40.91 & 1.71 & 70.79 & 1.97 & 56.90 & 1.99 & {\color{red} \textbf{78.92}} & 1.73 & {\color{blue} \textit{76.83}} & \textbf{0.74} \\
(CIFAR-10) Symmetric-80\% & 15.85 & 18.99 & 15.29 & 1.82 & {\color{blue} \textit{26.54}} & 1.98 & 23.50 & 2.00 & 25.51 & 1.73 & {\color{red} \textbf{40.24}} & \textbf{0.53} \\
(CIFAR-10) Asymmetric-40\% & 69.23 & 69.64 & 69.10 & 1.51 & {\color{blue} \textit{73.59}} & 1.99 & 68.45 & 2.00 & {\color{red} \textbf{76.13}} & 1.74 & 73.43 & \textbf{1.04} \\
\hline
(CIFAR-100) Symmetric-20\% & 35.51 & 36.04 & 33.82 & 1.75 & 44.03 & 1.97 & 49.24 & 1.99 & {\color{red} \textbf{53.10}} & 1.73 & {\color{blue} \textit{50.74}} & \textbf{0.93} \\
(CIFAR-100) Symmetric-50\% & 17.31 & 21.14 & 15.81 & 1.82 & 34.96 & 1.97 & 40.26 & 2.00 & {\color{red} \textbf{43.28}} & 1.73 & {\color{blue} \textit{40.98}} & \textbf{0.68} \\
(CIFAR-100) Symmetric-80\% & 4.25 & 7.48 & 4.03 & 1.90 & {\color{blue} \textit{14.81}} & 1.97 & 13.99 & 2.00 & 12.90 & 1.72 & {\color{red} \textbf{16.03}} & \textbf{0.52} \\
(CIFAR-100) Asymmetric-40\% & 27.91 & 27.11 & 26.95 & 1.75 & 28.69 & 1.98 & {\color{blue} \textit{34.30}} & 2.01 & 32.39 & 1.74 & {\color{red} \textbf{35.23}} & \textbf{0.98} \\
\hline
(Clothing1M) Best & 67.62 & 67.34 & 68.32 & \multirow{2}{*}{1.97} & 68.37 & \multirow{2}{*}{1.99} & 68.51 & \multirow{2}{*}{1.99} & {\color{red} \textbf{70.30}}* & \multirow{2}{*}{2.00} & {\color{blue} \textit{70.28}} & \multirow{2}{*}{\textbf{0.41}} \\
(Clothing1M) Last & 66.05 & 66.73 & 67.69 & & 68.12 & & 68.51 & & {\color{blue} \textit{69.79}}* & & {\color{red} \textbf{70.28}} & \\
\hline
\end{tabular}
\end{small}
\vskip -3mm
\begin{footnotesize}
\begin{tablenotes}
      \item{$\cdot$} Acc: Test accuracy (\%) averaged over five trials. The best test accuracy is indicated with {\color{red} \textbf{red bold}} and the second is indicated with {\color{blue} \textit{blue italic}}. 
      \item{$\cdot$} $\times$T: Computation time increase from Standard. Computation time of F-correction is the same as Standard (i.e., $\times$T=1) and thus omitted.
     \item{$\cdot$} For Clothing1M, we showed the test accuracy in the epoch with the highest validation accuracy (Best) and in the last epoch (Last).
     \item{$\cdot$} (*) For JoCoR, we refer to the accuracy on Clothing1M in~\cite{wei2020combating} because the hyperparameters were not suggested.
    \end{tablenotes}
\end{footnotesize}
\end{threeparttable}
\vskip -3mm
\end{table*}
\begin{table*}[t]
\centering
\begin{threeparttable}
\bigskip
\caption{Average label precision (\%) ( clean samples / selected samples) over the last 10 epochs.\vspace{-2mm}}
\label{table:precision}
\begin{small}
\begin{tabular}{c|c|c|c|c|c|c|c|c|c|c|c|c|c|c|c}
\hline
\multirow{2}{*}{Noise} & \multicolumn{3}{c|}{Decouple} & \multicolumn{3}{c|}{Co-teaching} & \multicolumn{3}{c|}{Co-teaching+} & \multicolumn{3}{c|}{JoCoR} & \multicolumn{3}{c}{\Ours(Ours)} \\
  & M & C10 & C100 & M & C10 & C100 & M & C10 & C100 & M & C10 & C100 & M & C10 & C100 \\
\hline\hline
S20 & 37.31 & 68.19 & 72.50 & 95.38 & 92.20 & 91.87 & 79.93 & 56.95 & 48.73 & {\color{blue} \textit{98.14}} & {\color{blue} \textit{96.83}} & {\color{blue} \textit{95.91}} & {\color{red} \textbf{99.72}} & {\color{red} \textbf{98.16}} & {\color{red} \textbf{97.65}} \\
S50 & 31.83 & 40.44 & 43.37 & 89.78 & 82.65 & 80.11 & 49.32 & 5.64 & 25.34 & {\color{blue} \textit{95.43}} & {\color{blue} \textit{91.04}} & {\color{blue} \textit{87.40}} & {\color{red} \textbf{99.66}} & {\color{red} \textbf{93.16}} & {\color{red} \textbf{91.24}} \\
S80 & 16.91 & 18.19 & 18.24 & 77.48 & {\color{blue} \textit{36.41}} & {\color{blue} \textit{48.71}} & 12.08 & 19.73 & 18.83 & {\color{blue} \textit{88.19}} & 35.95 & 45.15 & {\color{red} \textbf{97.19}} & {\color{red} \textbf{51.43}} & {\color{red} \textbf{51.88}} \\
A40 & 52.65 & 68.57 & 56.91 & 93.52 & 86.79 & 62.15 & 79.97 & 79.58 & 36.87 & {\color{blue} \textit{95.99}} & {\color{red} \textbf{87.94}} & {\color{blue} \textit{63.59}} & {\color{red} \textbf{99.11}} & {\color{blue} \textit{87.47}} & {\color{red} \textbf{68.41}} \\
\hline
\end{tabular}
\end{small}
\vskip -3mm
\begin{footnotesize}
\begin{tablenotes}
      \item{$\cdot$} Clothing1M has no GT about noisy labels and thus it is impossible to show its label precision.
      \item{$\cdot$} M:MNIST, C10: CIFAR-10, C100: CIFAR-100. \quad {$\cdot$} S\{20,50,80\}: Symmetric-\{20,50,80\}\%. A40: Asymmetric-40\%.
    \end{tablenotes}
\end{footnotesize}
\end{threeparttable}\vspace{-3mm}
\end{table*}
\addtolength{\tabcolsep}{2pt}
\subsection{Selection risk bound of \Ours}
Based on Theorem~\ref{theo:rb}, which is a result for the general $k$-set selection problem, we now derive a new theoretical result specific to \Ours.
Specifically, this theoretical result proves that \Ours select samples with a smaller noise-risk on average over $t=1$ to $T$ and therefore will give a more direct interpretation of how theoretical support of \Ours is beneficial for the sample selection. We define the average selection risk as $\ASE_{(\bd_1,\ldots, \bd_T)} = (1/T)\sum_{t=1}^T \bd_t \cdot \btheta_t$ for the length $T$ of sequences of selected samples $\bd_1,\ldots, \bd_T$ and noise-level vectors $\btheta_1,\ldots, \btheta_T$.
\par
\begin{coro} For a sequence of noise-risk vectors $\btheta_1, \ldots, \btheta_T$, we assume that \Ours selects $\bd_1, \ldots, \bd_T$ samples and the best $k$-set sample selection $\bestd = \arg\min_{\bd \in \mathcal{D}}\sum_{t=1}^T \bd \cdot \btheta_t$ achieves $\alpha k$ average selection risk (i.e., $\ASE_{(\bd^*,\ldots, \bd^*)}=\alpha k$), where $0 \leq \alpha \leq 1$. Then, $\ASE_{(\bd_1, \ldots, \bd_T)}$  is upper bounded as follows:
  \[
  \mathbb{E}[\ASE_{(\bd_1, \ldots, \bd_T)}] \leq \alpha k \left(\frac{2\sqrt{2 \ln n }}{\sqrt{T\alpha}} + 1\right)
  \]
\end{coro}
%
\noindent
The proof is shown in supplementary materials.\par
The parenthesized term 
is more than $1$ and it becomes closer to $1$ when 
the fraction part
is smaller. In other words, $\ASE_{(\bd_1, \ldots, \bd_T)}$ becomes $\alpha k$ when 
the fraction part 
is sufficiently small.
The fraction part $(2\sqrt{2 \ln n }/{\sqrt{T\alpha}})$ is not so large when we use a large training sample set because the dependency on the sample size $n$ is logarithmic, and thus that is a great advantage for the sample selection for DNN training which basically requires large training samples.
Moreover, the fraction part becomes smaller with increasing $T$. Consequently, $\ASE_{(\bd_1, \ldots, \bd_T)}$ will get close to $\alpha k$ along with $t$.
Since $\ASE_{(\bd_1, \ldots, \bd_T)}$ and $\alpha k$ are the average selection risks of \Ours and $\bd^*$ (i.e., the selection method that knows $\bd^*$ as an oracle), this conclusion proves that the DNN on TAkS can be trained with samples $\bd_t$ with as small noise-risk as $\bestd$.


\section{Experimental Results\label{sec:experiment}}
\subsection{Noise-risk assessment}
\label{subsec:noise-risk-vector}
In this paper, we assess a noise-risk of $x_i$ as follows:
\begin{align}
\label{align:loss}
\theta_{i,t} = \frac{1- \mathrm{is}(f_t(x_i)= y_i)p(f_t(x_i))}{2},
\end{align}
where
$f_t(x_i)$ is the predicted label of $x_i$.
The function $\mathrm{is}(\cdot)$ returns $+1$ if $\cdot$ is true and returns $-1$ otherwise, and $p$ denotes the probability (i.e., softmax output) for the predicted label $f_t(x_i)$.
Eq.~(\ref{align:loss}) is motivated on the following hypothesis:
if the sample $x_i$ is predicted as the class $y_i$ with higher confidence, it should have a correct label, and if the sample is predicted as a class other than $y_i$ with higher confidence, it should have an incorrect label.
\par
The above noise-risk assessment is reasonable because of the following reasons. First, selecting samples based on the confidence has been commonly used in several existing sample selection methods~(e.g., \cite{han2018co,wei2020combating}).
Second, it is known that DNN basically fits easy samples first~\cite{arpit2017closer}, and that suggests that clean (and easy) samples have high confidence even in the earlier epochs.
Finally, with Eq.~(\ref{align:loss})  incorrectly labeled samples are basically difficult to learn, i.e., DNN is hard to increase the confidence of the incorrect class label. 
The preliminary experiment verifies the above, and thus the noise-risk assessment is reasonable, as shown in supplementary materials.
\subsection{Experimental setup}
\label{sec:exp_setup}
\noindent{\bf Datasets}\quad
We follow the evaluation scenario of the state-of-the-art trials for noisy-labeled data. We, therefore, used four datasets, MNIST~\cite{lecun2010mnist}, CIFAR-10, CIFAR-100~\cite{krizhevsky2009learning}, and Clothing1M~\cite{xiao2015learning} and applied symmetric noise~\cite{van2015learning} and asymmetric noise~\cite{patrini2017making} to those datasets to the first three datasets, which have only clean labels.
For symmetric noise, we randomly choose a certain percentage (20\%, 50\%, and 80\%) of the training samples and replaced their original (i.e., clean) label with one of the other (i.e., incorrect) labels.
For asymmetric noise, we replaced the original label with the label of a prespecified confusing class. Note that 40\% of asymmetric noise actually means that 20\% of the labels are flipped, because the noise is applied to only half of the classes in \cite{wei2020combating}. \par
Clothing1M is a large dataset and widely used to evaluate the methods for learning with noisy labels~\cite{li2019learning,yi2019probabilistic,wei2020combating}. It has no ground-truth about noisy labels, but its noise rate is estimated at about 40\%~\cite{xiao2015learning,yi2019probabilistic}. For preprocessing, we resized the image samples to 256$\times$256, cropped them at the center to size 224$\times$224, and normalized them.
\par
\begin{figure*}[t]
\centering
\includegraphics[width=1\linewidth]{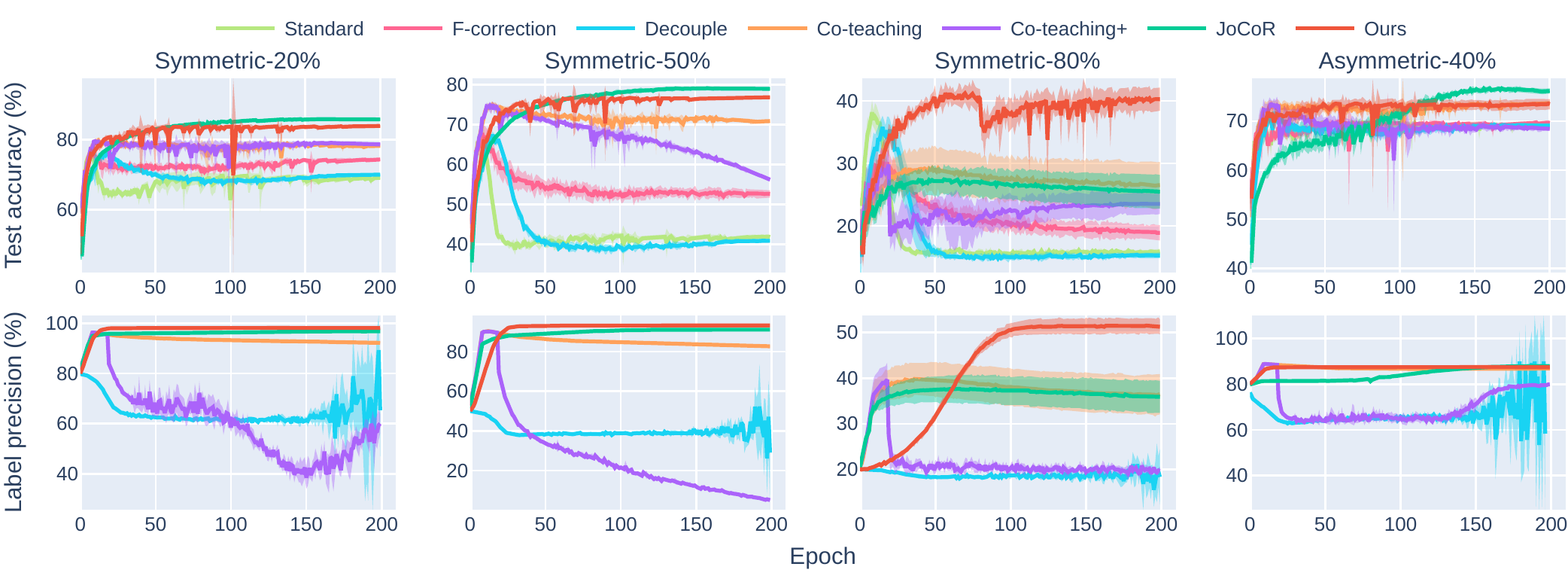}\\[-5mm]
\caption{Test accuracy (Top) and label precision (Bottom) graphs with error band on CIFAR-10.}
\label{fig:cifar10_graph}
\vskip -5mm
\end{figure*}
\noindent{\bf Baselines}\quad
Following~\cite{wei2020combating}, we compared \Ours with five state-of-the-art sample selection methods: ``F-correction''~\cite{patrini2017making}, ``Decouple''~\cite{malach2017decoupling}, ``Co-teaching''~\cite{han2018co}, ``Co-teaching+''~\cite{yu2019does}, and ``JoCoR''~\cite{wei2020combating}. We also compared \Ours with the standard training strategy (``Standard'') which uses the whole (noisy) training set.
\par
\noindent{\bf Evaluation metrics}\quad
We measured the following three metrics; test accuracy, label precision, and training time. Test accuracy represents the proportion of correctly classified samples among the test samples, by the trained DNN. Following~\cite{han2018co,wei2020combating}, we focused on average test accuracy over the last 10 epochs (of $T=200$ epochs). Label precision shows the percentage of clean samples among the selected samples; higher label precision means more clean samples are used for training a DNN. Training time is the time taken to train a DNN. For F-correction, we exclude its (long) preprocessing time to estimate the label transition matrix for a fair comparison; consequently, its computation time becomes the same as Standard. \par
We repeated the experiments five times for all datasets except Clothing1M and calculated the average for each metric. For Clothing1M, we conducted an experiment once, and label precision is not given because it has no ground-truth about noisy labels.
\par
\noindent{\bf DNN structure and optimizer}\quad
For a fair comparison, we follow the setup of JoCoR~\cite{wei2020combating}.
More details can be found in supplementary materials.
\par
\noindent{\bf Hyperparameter search}\quad
To fix the hyperparameters $\eta$ and $k$, we split the noisy training samples into 80\% noisy training samples and 20\% {\em noisy validation samples}. With the validation samples, the best $\eta$ was selected from its candidates $\eta\in\{10^{-4}, 5\times 10^{-4}, 10^{-3}, 5\times 10^{-3}\} \times \sqrt{kT}$. \par

The hyperparameter $k$ should be fixed by using noise rate estimation. 
Although there are several techniques to estimate the noise rate~\cite{liu2015classification,yu2018efficient}, it is still hard to estimate the true noise rate $\gamma$. Therefore, assuming that we could roughly estimate the noise rate as $\gamma$, we choose the best $k$
from $\{1-\gamma-0.15, 1-\gamma-0.1, \ldots, 1-\gamma+0.15\} \times n$
with the noisy validation samples. \par
Note that the above conditions are different from the state-of-the-art attempts and {\em dis}advantageous to \Ours. In fact, Co-teaching, Co-teaching+, and JoCoR use the true noise rate directly~\cite{han2018co,yu2019does,wei2020combating}. Furthermore, JoCoR searched for hyperparameters using {\em clean validation samples}~\cite{wei2020combating}. We kept these conditions in our experiments. In contrast, hyperparameters of \Ours are searched by using the noisy validation samples and the rough estimation of $\gamma$. Although our hyperparameter search mimics a more practical scenario, it handicaps \Ours. The tuned hyperparameters are detailed in supplementary materials.
\par
\subsection{Experimental results}
\noindent{\bf Classification performance}\quad
Table~\ref{table:accuracy} shows the average test accuracy over the last 10 epochs. For Clothing1M, it shows the test accuracy when the trained model achieved the best validation accuracy as ``Best'' and the test accuracy at the end of training as ``Last'' (by following~\cite{wei2020combating}). \Ours achieved a noticeable performance improvement from Standard and, more importantly, outperformed all the other baselines, achieving the best or second-best test accuracy in most cases. In particular, \Ours achieved significantly higher test accuracy than all other baselines when the training samples contained many incorrectly labeled samples, such as Symmetric-80\%. 
\par
More precisely, \Ours achieved the highest test accuracy for all noise cases on MNIST. On CIFAR-10, \Ours and JoCoR performed better than the other baselines. However, in the Symmetric-80\% case, the test accuracy of all baselines, including JoCoR, was significantly degraded from Symmetric-20\% and 50\%. On the other hand, \Ours still maintained higher test accuracy. Recall that its theoretical supports hold even in severely noisy cases, and \Ours can keep its reasonable performance if the sequence of the noise-risk vectors is given correctly. For Clothing1M, \Ours achieved the best accuracy in both the ``Best'' and ``Last'' scenarios. It is noteworthy again that our hyperparameters were determined in a more challenging scenario than the baselines, where the exact noise ratio and/or a clean validation set were used.
\par
We also plotted the test accuracy curves on CIFAR-10 in Fig.~\ref{fig:cifar10_graph} (Curves for other datasets can be found in the supplementary materials; most of the discussion below also applies to these datasets.). In most cases, the accuracy of Standard, F-correction, and Decouple increased and reached a high peak at the beginning when they learned general features. After that, they started overfitting the incorrectly labeled training samples, and the test accuracy dropped rapidly. In contrast, the other baselines and \Ours maintained high test accuracy. A closer observation reveals that the test accuracy of \Ours steadily increased while the test accuracy of the baselines gradually decreased for difficult cases, such as Symmetric-80\%.
\par

\noindent{\bf Label precision}\quad
The robustness of \Ours can be explained with label precision. Table~\ref{table:precision} shows the average label precision over the last 10 epochs. In all noise cases of all datasets except for one, \Ours achieved the best label precision, and even in the one case, \Ours achieved the second-best. It means that \Ours avoided incorrectly labeled samples successfully and prevented the DNN from overfitting them. 
\par
Fig.~\ref{fig:cifar10_graph} plots the label precision on CIFAR-10 at each epoch. In all cases, the label precision of \Ours did not decrease and rather kept increasing or saturating. Although high label precision does not directly guarantee high test accuracy, it surely confirms the high noise tolerance of \Ours. It was especially noticeable when there was much noise in the training sample, such as the Symmetric-80\%. In the case, Fig.~\ref{fig:cifar10_graph} shows that the test accuracy of the baselines, including Co-teaching and JoCoR, began to deteriorate as their label precision began to decrease. In contrast, the accuracy of \Ours has steadily increased, and even longer training is expected to lead to better performance. 
\par
\begin{figure}[t]
\centering
\includegraphics[width=1\linewidth]{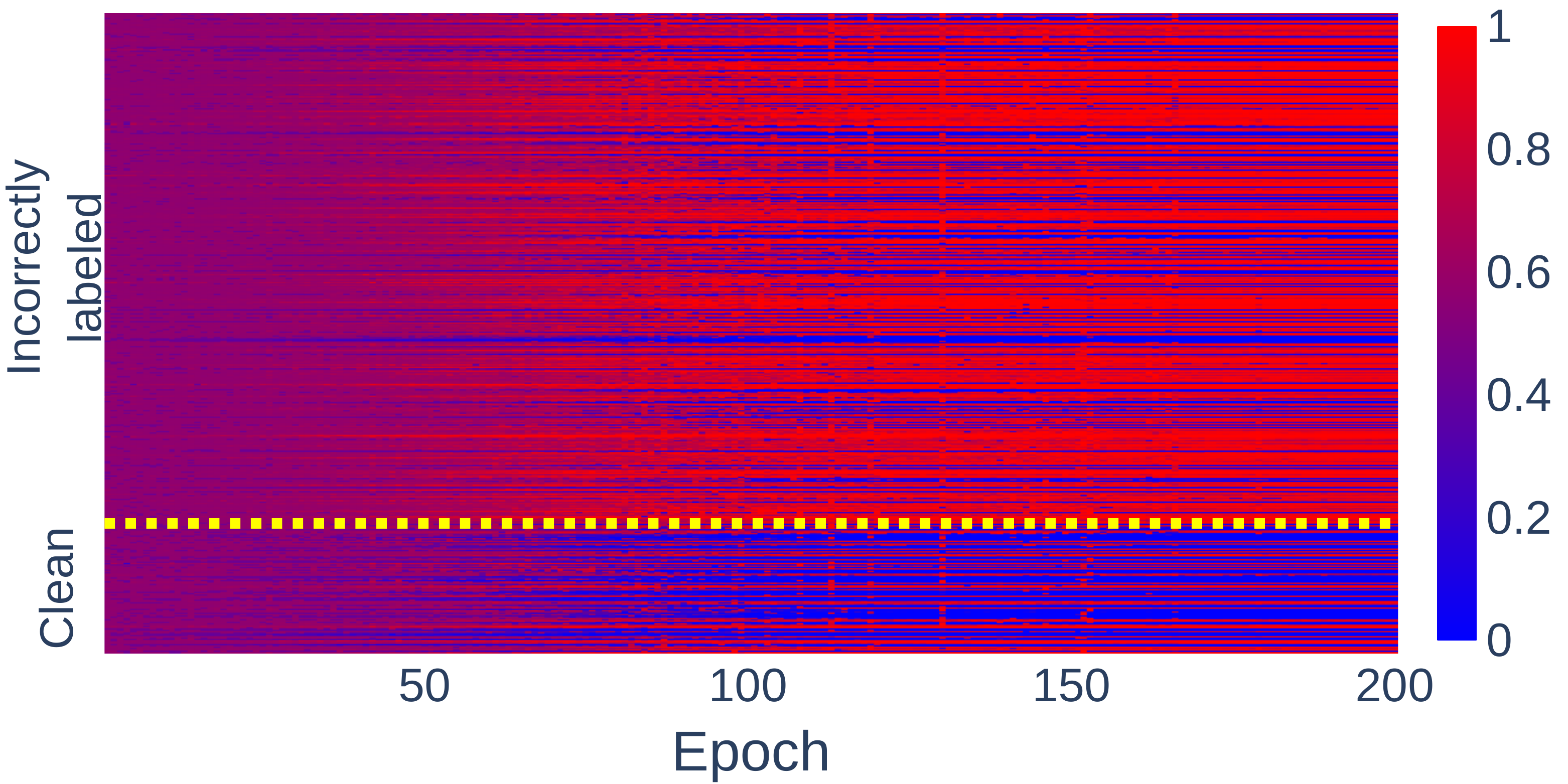}\\[-3mm]
\caption{Change of noise-risk $\theta_{i,t}$ on CIFAR-10 in Symmetric-80\%.}
\label{fig:cifar10_loss}\vspace{-3mm}
\end{figure}

Fig.~\ref{fig:cifar10_loss} shows the change in the noise-risk $\theta_{i,t}$ of each training sample on CIFAR-10 with Symmetric-80\% over the epochs. After the beginning epochs with almost uniform noise-risk values over all samples, most clean samples were properly distinguished by their small noise-risk values. 
\par
\begin{figure}[t]
\centering
\includegraphics[width=1\linewidth]{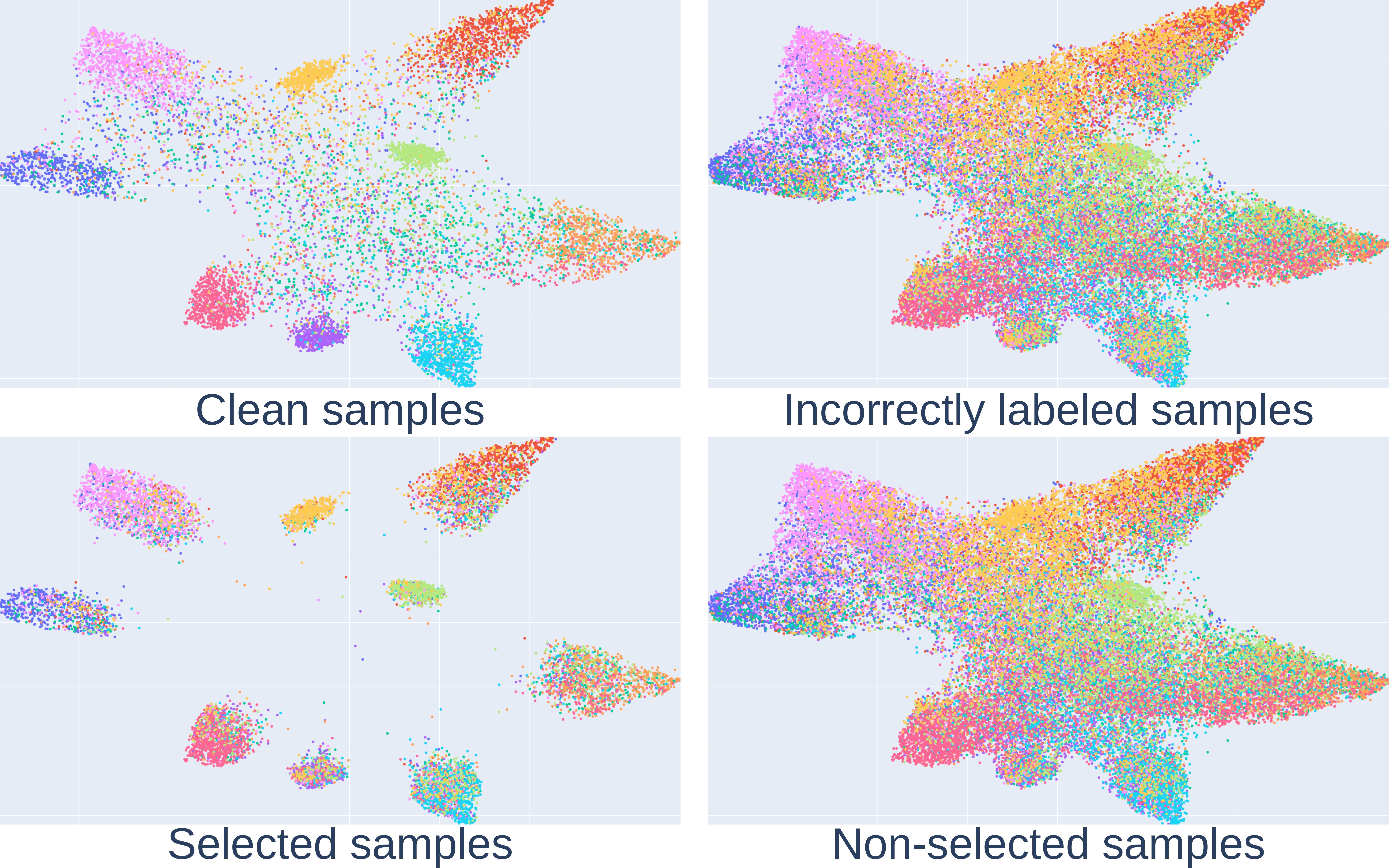}\\[-2mm]
\caption{Visualization of the feature distribution of the noisy training samples, CIFAR-10 in Symmetric-80\%, by \Ours. Colors represent classes. Incorrectly labeled samples are plotted with the color of their true class.}
\label{fig:cifar10_samples}\vspace{-3mm}
\end{figure}

Fig.~\ref{fig:cifar10_samples} shows the feature distributions in the Symmetric-80\% case on CIFAR-10 by \Ours. UMAP~\cite{mcinnes2018umap} was used for two-dimensional visualization. Clean training samples of the same class form a cluster, whereas incorrectly labeled training samples tended to be scattered. 
This allowed \Ours to classify the clean test samples accurately. Fig.~\ref{fig:cifar10_samples} also shows the feature distributions of the selected and the non-selected samples of \Ours. The samples belonging to the clusters for the clean samples were mainly selected, and most incorrectly-labeled samples outside of the clusters were not selected. 
This suggests that the noise-risk vectors worked successfully to exclude suspicious samples that were far from the cluster of clean samples.
\par

\noindent{\bf Computation time}\quad
It should be emphasized that \Ours is more efficient than the baselines. Table~\ref{table:accuracy} shows the training time of each method and proves that \Ours was faster than the baselines in almost all cases. Moreover, \Ours was even faster than Standard in most cases, because \Ours uses only $k<n$ samples, whereas Standard uses all $n$ samples. For example, $k$ became the smallest for Symmetric-80\%, and thus, \Ours became twice and four times as fast as Standard and co-training methods, respectively.
\par

\subsection{Ablation study}
To prove the effectiveness of FPL for a smaller total selection risk,
we conduct an ablation study with the same settings as Sec.~\ref{sec:exp_setup}. 
More specifically, we compare FPL with the following two algorithms, ``Greedy'' and ``Naive.''
Greedy does not focus on \emph{total} selection risk. Instead, it selects $\bd_{t}$ which minimizes $\bd_t \cdot \btheta_{t-1}$ at every epoch $t$. This means Greedy trusts temporary noise-risk vector completely. 
Naive selects $\bd_{t}$ which minimizes $\sum_{\tau=1}^t \bd_t \cdot \btheta_{\tau-1}$ at epoch $t$. That is, it selects the $k$-set that gives the current minimum total selection risk without the perturbation by $\boldsymbol{r}$. Note that Naive corresponds to FTL~\cite{hazan2016introduction}.
Table~\ref{table:ablation} shows the average test accuracy of the above three methods over the last 10 epochs. 
By comparing between Tables~\ref{table:accuracy} and \ref{table:ablation}, 
it is proved that \Ours outperforms the others. Especially,  
Greedy performed significantly worse than ours for high noise-rate data, because the noise-risk vector can be more unstable for noisier data.
Naive fails in many cases because it has suffered from the unreliable noise-risk in early epochs. Ours can overcome these problems by the perturbation mechanism with the support of the lower regret bound. 
\addtolength{\tabcolsep}{-1pt}
\begin{table}[t]
\centering
\begin{threeparttable}
\caption{Results of ablation study.\vspace{-2mm}}
\label{table:ablation}
\begin{small}
\begin{tabular}{c|c|c|c|c|c|c}
\hline
\multirow{2}{*}{Noise} & \multicolumn{3}{c|}{Greedy} & \multicolumn{3}{c}{Naive} \\
  & M & C10 & C100 & M & C10 & C100 \\
\hline\hline
S20 & \lowacc{91.78} & \lowacc{83.64} & \lowacc{49.29} & \lowacc{92.62} & \lowacc{82.07} & \lowacc{45.11} \\
S50 & \lowacc{88.22} & 76.96 & \lowacc{38.08} & \lowacc{89.28} & \lowacc{66.65} & \lowacc{21.85} \\
S80 & \lowacc{70.22} & \lowacc{25.09} & \lowacc{11.33} & \lowacc{66.06} & \lowacc{20.16} & \lowacc{4.20} \\
A40 & \lowacc{92.24} & \lowacc{71.65} & \lowacc{33.26} & \lowacc{89.84} & \lowacc{69.12} & \lowacc{33.10} \\
\hline
\end{tabular}
\end{small}
\vskip -3mm
\begin{footnotesize}
\begin{tablenotes}
      \item{$\cdot$} Gray cell indicates that it is less accurate than \Ours.
      \item{$\cdot$} See Table~\ref{table:precision} for the notations, such as M and S20.
\end{tablenotes}
\end{footnotesize}
\end{threeparttable}
\vspace{-3mm}
\end{table}

\section{Conclusion}
We tackled the learning task from noisy-labeled data and proposed a novel DNN training method, \Ours, based on adaptive sample selection. The sample selection algorithm was inspired by an adaptive $k$-set selection, which allows \Ours to have strong theoretical support. In the experiments, we showed that the proposed method worked more efficiently and effectively than state-of-the-art methods. We observed the behaviors of the proposed method and confirmed that it could train DNNs while excluding the incorrectly-labeled samples appropriately.

{\small
\bibliographystyle{ieee_fullname}
\bibliography{egbib}

\begin{thebibliography}{10}\itemsep=-1pt

\bibitem{arpit2017closer}
Devansh Arpit, Stanislaw~K Jastrzebski, Nicolas Ballas, David Krueger, Emmanuel
  Bengio, Maxinder~S Kanwal, Tegan Maharaj, Asja Fischer, Aaron~C Courville,
  Yoshua Bengio, et~al.
\newblock A closer look at memorization in deep networks.
\newblock In {\em Proc. ICML}, 2017.

\bibitem{beyer2020we}
Lucas Beyer, Olivier~J H{\'e}naff, Alexander Kolesnikov, Xiaohua Zhai, and
  A{\"a}ron van~den Oord.
\newblock Are we done with imagenet?
\newblock {\em arXiv preprint arXiv:2006.07159}, 2020.

\bibitem{breunig2000lof}
Markus~M Breunig, Hans-Peter Kriegel, Raymond~T Ng, and J{\"o}rg Sander.
\newblock Lof: identifying density-based local outliers.
\newblock In {\em Proc. SIGMOD}, 2000.

\bibitem{cohen2015following}
Alon Cohen and Tamir Hazan.
\newblock Following the perturbed leader for online structured learning.
\newblock In {\em Proc. ICML}, 2015.

\bibitem{deng2009imagenet}
Jia Deng, Wei Dong, Richard Socher, Li-Jia Li, Kai Li, and Li Fei-Fei.
\newblock Imagenet: A large-scale hierarchical image database.
\newblock In {\em Proc. CVPR}, 2009.

\bibitem{frenay2013classification}
Beno{\^\i}t Fr{\'e}nay and Michel Verleysen.
\newblock Classification in the presence of label noise: a survey.
\newblock {\em TNNLS}, 25(5):845--869, 2013.

\bibitem{han2018co}
Bo Han, Quanming Yao, Xingrui Yu, Gang Niu, Miao Xu, Weihua Hu, Ivor Tsang, and
  Masashi Sugiyama.
\newblock Co-teaching: Robust training of deep neural networks with extremely
  noisy labels.
\newblock In {\em Proc. NeurIPS}, 2018.

\bibitem{hazan2016introduction}
Elad Hazan.
\newblock Introduction to online convex optimization.
\newblock {\em Foundations and Trends in Optimization}, 2(3-4):157--325, 2016.

\bibitem{he2016deep}
Kaiming He, Xiangyu Zhang, Shaoqing Ren, and Jian Sun.
\newblock Deep residual learning for image recognition.
\newblock In {\em Proc. CVPR}, 2016.

\bibitem{jiang2018mentornet}
Lu Jiang, Zhengyuan Zhou, Thomas Leung, Li-Jia Li, and Li Fei-Fei.
\newblock Mentornet: Learning data-driven curriculum for very deep neural
  networks on corrupted labels.
\newblock In {\em Proc. ICML}, 2018.

\bibitem{kalai2002geometric}
Adam Kalai and Santosh Vempala.
\newblock Geometric algorithms for online optimization.
\newblock {\em Journal of Computer and System Sciences}, 71(3):291--307, 2002.

\bibitem{kingma2014adam}
Diederik~P Kingma and Jimmy Ba.
\newblock Adam: A method for stochastic optimization.
\newblock In {\em Proc. ICLR}, 2015.

\bibitem{koolen2010hedging}
Wouter~M Koolen, Manfred~K Warmuth, and Jyrki Kivinen.
\newblock Hedging structured concepts.
\newblock In {\em Proc. COLT}, 2010.

\bibitem{krizhevsky2009learning}
Alex Krizhevsky and Geoffrey Hinton.
\newblock Learning multiple layers of features from tiny images.
\newblock 2009.

\bibitem{lecun2010mnist}
Yann LeCun, Corinna Cortes, and CJ Burges.
\newblock Mnist handwritten digit database.
\newblock 2010.

\bibitem{li2019learning}
Junnan Li, Yongkang Wong, Qi Zhao, and Mohan~S Kankanhalli.
\newblock Learning to learn from noisy labeled data.
\newblock In {\em Proc. CVPR}, 2019.

\bibitem{li2017learning}
Yuncheng Li, Jianchao Yang, Yale Song, Liangliang Cao, Jiebo Luo, and Li-Jia
  Li.
\newblock Learning from noisy labels with distillation.
\newblock In {\em Proc. CVPR}, 2017.

\bibitem{liu2015classification}
Tongliang Liu and Dacheng Tao.
\newblock Classification with noisy labels by importance reweighting.
\newblock {\em TPAMI}, 38(3):447--461, 2015.

\bibitem{malach2017decoupling}
Eran Malach and Shai Shalev-Shwartz.
\newblock Decoupling ``when to update'' from ``how to update''.
\newblock In {\em Proc. NeurIPS}, 2017.

\bibitem{mcinnes2018umap}
Leland McInnes, John Healy, and James Melville.
\newblock Umap: Uniform manifold approximation and projection for dimension
  reduction.
\newblock {\em arXiv preprint arXiv:1802.03426}, 2018.

\bibitem{menon2015learning}
Aditya Menon, Brendan Van~Rooyen, Cheng~Soon Ong, and Bob Williamson.
\newblock Learning from corrupted binary labels via class-probability
  estimation.
\newblock In {\em Proc. ICML}, 2015.

\bibitem{musser1997introspective}
David~R Musser.
\newblock Introspective sorting and selection algorithms.
\newblock {\em Software: Practice and Experience}, 27(8):983--993, 1997.

\bibitem{natarajan2013learning}
Nagarajan Natarajan, Inderjit~S Dhillon, Pradeep~K Ravikumar, and Ambuj Tewari.
\newblock Learning with noisy labels.
\newblock In {\em Proc. NeurIPS}, 2013.

\bibitem{patrini2017making}
Giorgio Patrini, Alessandro Rozza, Aditya Krishna~Menon, Richard Nock, and
  Lizhen Qu.
\newblock Making deep neural networks robust to label noise: A loss correction
  approach.
\newblock In {\em Proc. CVPR}, 2017.

\bibitem{ren2018learning}
Mengye Ren, Wenyuan Zeng, Bin Yang, and Raquel Urtasun.
\newblock Learning to reweight examples for robust deep learning.
\newblock In {\em Proc. ICML}, 2018.

\bibitem{rodrigues2018deep}
Filipe Rodrigues and Francisco~Camara Pereira.
\newblock Deep learning from crowds.
\newblock In {\em Proc. AAAI}, 2018.

\bibitem{shen2019learning}
Yanyao Shen and Sujay Sanghavi.
\newblock Learning with bad training data via iterative trimmed loss
  minimization.
\newblock In {\em Proc. ICML}, 2019.

\bibitem{suehiro2012online}
Daiki Suehiro, Kohei Hatano, Shuji Kijima, Eiji Takimoto, and Kiyohito Nagano.
\newblock Online prediction under submodular constraints.
\newblock In {\em Proc. ALT}, 2012.

\bibitem{sukhbaatar2015training}
Sainbayar Sukhbaatar, Joan~Bruna Estrach, Manohar Paluri, Lubomir Bourdev, and
  Robert Fergus.
\newblock Training convolutional networks with noisy labels.
\newblock In {\em Proc. ICLR}, 2015.

\bibitem{van2015learning}
Brendan Van~Rooyen, Aditya Menon, and Robert~C Williamson.
\newblock Learning with symmetric label noise: The importance of being
  unhinged.
\newblock In {\em Proc. NeurIPS}, 2015.

\bibitem{veit2017learning}
Andreas Veit, Neil Alldrin, Gal Chechik, Ivan Krasin, Abhinav Gupta, and Serge
  Belongie.
\newblock Learning from noisy large-scale datasets with minimal supervision.
\newblock In {\em Proc. CVPR}, 2017.

\bibitem{wang2018iterative}
Yisen Wang, Weiyang Liu, Xingjun Ma, James Bailey, Hongyuan Zha, Le Song, and
  Shu-Tao Xia.
\newblock Iterative learning with open-set noisy labels.
\newblock In {\em Proc. CVPR}, 2018.

\bibitem{warmuth2008randomized}
Manfred~K Warmuth and Dima Kuzmin.
\newblock Randomized online {PCA} algorithms with regret bounds that are
  logarithmic in the dimension.
\newblock {\em JMLR}, 9(Oct):2287--2320, 2008.

\bibitem{wei2020combating}
Hongxin Wei, Lei Feng, Xiangyu Chen, and Bo An.
\newblock Combating noisy labels by agreement: A joint training method with
  co-regularization.
\newblock In {\em Proc. CVPR}, 2020.

\bibitem{xia2019anchor}
Xiaobo Xia, Tongliang Liu, Nannan Wang, Bo Han, Chen Gong, Gang Niu, and
  Masashi Sugiyama.
\newblock Are anchor points really indispensable in label-noise learning?
\newblock In {\em Proc. NeurIPS}, 2019.

\bibitem{xiao2015learning}
Tong Xiao, Tian Xia, Yi Yang, Chang Huang, and Xiaogang Wang.
\newblock Learning from massive noisy labeled data for image classification.
\newblock In {\em Proc. CVPR}, 2015.

\bibitem{yan2014learning}
Yan Yan, R{\'o}mer Rosales, Glenn Fung, Ramanathan Subramanian, and Jennifer
  Dy.
\newblock Learning from multiple annotators with varying expertise.
\newblock {\em Machine learning}, 95(3):291--327, 2014.

\bibitem{yao2020searching}
Quanming Yao, Hansi Yang, Bo Han, Gang Niu, and James Tin-Yau Kwok.
\newblock Searching to exploit memorization effect in learning with noisy
  labels.
\newblock In {\em Proc. ICML}, 2020.

\bibitem{yi2019probabilistic}
Kun Yi and Jianxin Wu.
\newblock Probabilistic end-to-end noise correction for learning with noisy
  labels.
\newblock In {\em Proc. CVPR}, 2019.

\bibitem{yu2019does}
Xingrui Yu, Bo Han, Jiangchao Yao, Gang Niu, Ivor~W Tsang, and Masashi
  Sugiyama.
\newblock How does disagreement help generalization against label corruption?
\newblock In {\em Proc. ICML}, 2019.

\bibitem{yu2018efficient}
Xiyu Yu, Tongliang Liu, Mingming Gong, Kayhan Batmanghelich, and Dacheng Tao.
\newblock An efficient and provable approach for mixture proportion estimation
  using linear independence assumption.
\newblock In {\em Proc. CVPR}, 2018.

\bibitem{yu2018learning}
Xiyu Yu, Tongliang Liu, Mingming Gong, and Dacheng Tao.
\newblock Learning with biased complementary labels.
\newblock In {\em Proc. ECCV}, 2018.

\bibitem{zhang2016understanding}
Chiyuan Zhang, Samy Bengio, Moritz Hardt, Benjamin Recht, and Oriol Vinyals.
\newblock Understanding deep learning requires rethinking generalization.
\newblock In {\em Proc. ICLR}, 2017.

\end{thebibliography}
}

\newpage
\clearpage
\appendix
\section{Proof of Corollary~2}


\begin{proof}
 For brevity, we simply use   $R_T$ as $\mathbb{E}[R_T]$. We first note that $kT$ in Theorem~1 is used as a trivial upper bound of
  the total selection risk of the best $k$-set selection
  (i.e., $\sum_{t=1}^T\bd^* \cdot \btheta_t \leq kT$ for any $\bd^*$ and $\btheta_1, \ldots, \btheta_T$).
  By the assumption that the best $k$-set selection achieves $\alpha k$ selection risk, we replace $kT$ with the actual total selection risk of the best selection $\alpha k T$.
  Thus, the regret bound is rewritten as follows:
  \[
   R_T \leq 2\sqrt{2\alpha kT \ln \binom{n}{k}}.
   \]
   Applying $\binom{n}{k} \leq n^k$ for simplicity, we have the following:
   \begin{align}
     \label{align:other_bound}
   R_T \leq 2 \sqrt{2\alpha kT \ln n^k} = 2k \sqrt{2\alpha T \ln n}.
   \end{align}
   By the definition of regret, the average selection error in \Ours
   is written as follows:
   \begin{align*}
     \frac{1}{T} \sum_{t=1}^T \bd_t \cdot \btheta_t &= \frac{R_T}{T} + \frac{\sum_{t=1}^T\bd^* \cdot \btheta_t}{T}=\frac{R_T}{T} + \alpha k
   \end{align*}
   Applying the bound~(\ref{align:other_bound}), we have the result:
   \begin{align*}
     \frac{1}{T} \sum_{t=1}^T \bd_t \cdot \btheta_t&\!\leq\!
     \frac{2k\sqrt{2\alpha T\ln n}}{T}\!+\!\alpha k\!=\! \alpha k\!\left(\!\frac{2 \sqrt{2 \ln n}}{\sqrt{T \alpha}}\!+\!1\!\right). 
   \end{align*}
   Finally we take into account the randomness of FPL, we obtain the the target bound.
\end{proof}

\section{Experimental details}
\subsection{Implementation}
We implemented \Ours and baselines with PyTorch and the code of our implementation can be found from attachment. All experiments were conducted on NVIDIA 1080Ti.

\subsection{Datasets}
The detail of the datasets used in our experiments is shown in Table~\ref{table:dataset}.

\subsection{DNN structure and optimizer}
The DNN architectures used on MNIST, CIFAR-10, and CIFAR-100 are shown in Table~\ref{table:dnn}. We used a MLP with one hidden layer for MNIST, a CNN with six hidden layers for CIFAR-10 and CIFAR-100, and ResNet-18~\cite{he2016deep} for Clothing1M. To optimize the DNN, we used Adam optimizer~\cite{kingma2014adam} ($\beta_1=0.9$ and $\beta_2=0.999$). For MNIST, CIFAR-10, and CIFAR-100, a learning rate was set to $10^{-3}$ up to 80 epochs and linearly decreased to $T=200$ epochs to become 0 after that. For Clothing1M, the learning rate was set for each of five epochs as $8\times 10^{-4}, 5\times 10^{-4}$, and $5\times 10^{-5}$ during 15 epochs. More details can be found in our code.

\begin{table}[t]
\centering
\begin{threeparttable}
\caption{The detail of datasets.\vspace{-2mm}}
\label{table:dataset}
\begin{tabular}{|c|c|c|c|}
\hline
           & \#training & \#test & \#class \\ \hline
MNIST      & 60000      & 10000  & 10      \\ \hline
CIFAR-10   & 50000      & 10000  & 10      \\ \hline
CIFAR-100  & 50000      & 10000  & 100     \\ \hline
Clothing1M & 1000000    & 10526  & 14      \\ \hline
\end{tabular}
\end{threeparttable}
\end{table}

\begin{table}[t]
\centering
\begin{threeparttable}
\caption{The detail of DNN architectures.\vspace{-2mm}}
\label{table:dnn}
\begin{tabular}{|c|c|c|}
\hline
Target                                                                  & MNIST                    & CIFAR-10 and CIFAR-100                                                               \\ \hline
Input                   & $28\times28$                    & $32\times32$                                                                              \\ \hline
\multirow{3}{*}{{\begin{tabular}[c]{@{}c@{}}Hidden\\ layer\end{tabular}}} & \multirow{3}{*}{fc, 256} & \begin{tabular}[c]{@{}c@{}}$3\times3$ conv, 64. BN\\ $3\times3$ conv, 64. BN\\ $2\times2$ max-pool\end{tabular}   \\ \cline{3-3} 
                        &                          & \begin{tabular}[c]{@{}c@{}}$3\times3$ conv, 128. BN\\ $3\times3$ conv, 128. BN\\ $2\times2$ max-pool\end{tabular} \\ \cline{3-3} 
                        &                          & \begin{tabular}[c]{@{}c@{}}$3\times3$ conv, 196. BN\\ $3\times3$ conv, 16. BN\\ $2\times2$ max-pool\end{tabular}  \\ \hline
Output                  & \multicolumn{2}{c|}{\begin{tabular}[c]{@{}c@{}}fc, \#class\\ softmax\end{tabular}}                              \\ \hline
\end{tabular}
\vskip -3mm
\begin{tablenotes}
      \item{$\cdot$} ReLU was used as an activation function.
\end{tablenotes}
\end{threeparttable}
\end{table}

\begin{table}[t]
\centering
\begin{threeparttable}
\caption{Tuned $\eta$ and $k$ for each dataset.\vspace{-2mm}}
\label{table:hyper}
\begin{small}
\begin{tabular}{c|c|c|c|c|c|c|c|c}
\hline
\multirow{2}{*}{} & \multicolumn{4}{c|}{$\eta$ ($\times \frac{\sqrt{kT}}{10000}$)} & \multicolumn{4}{c}{$k$ ($\times \frac{n}{100}$)} \\
  & M & C10 & C100 & CM & M & C10 & C100 & CM \\
\hline\hline
S20 & 5 & 5 & 5 & \multirow{4}{*}{10} & 65 & 75 & 65 & \multirow{4}{*}{50} \\
S50 & 50 & 10 & 10 & & 35 & 45 & 35 \\
S80 & 50 & 50 & 50 & & 15 & 20 & 15 \\
A40 & 10 & 5 & 5 & & 70 & 80 & 70 \\
\hline
\end{tabular}
\end{small}
\vskip -3mm
\begin{footnotesize}
\begin{tablenotes}
      \item{$\cdot$} See Table~\ref{table:hyper} for the notations, such as M and S20.
      \end{tablenotes}
\end{footnotesize}
\end{threeparttable}
\end{table}

\section{Noise-risk vector validation}
\begin{figure}[t]
\centering
\includegraphics[width=1\linewidth]{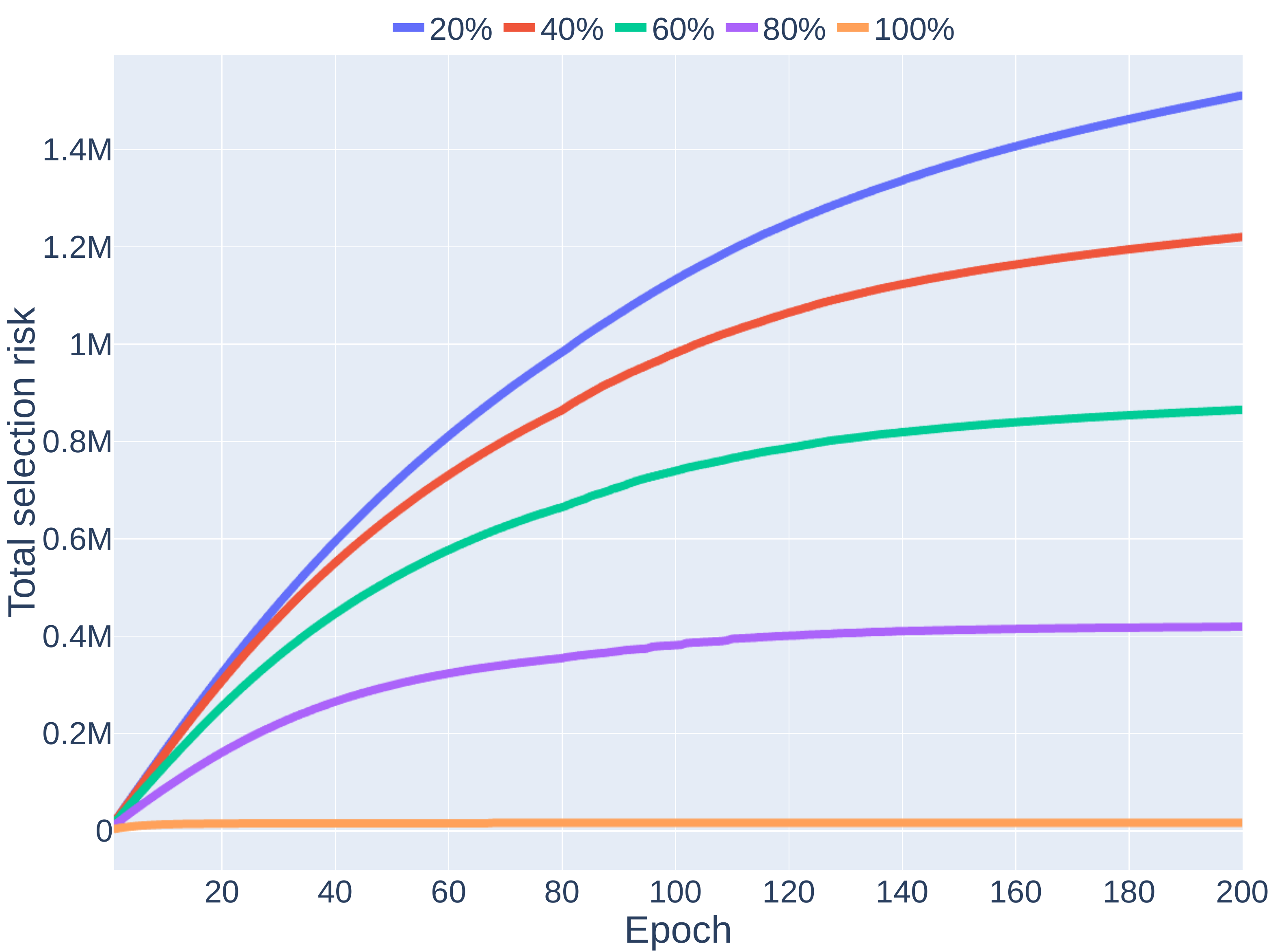}
\caption{Total selection risk of sample selection for different clean sample ratios.}
\label{fig:append_total_risk}
\end{figure}
The proposed noise-risk for $x_i$ is
\[
\theta_{i,t} = \frac{1- \mathrm{is}(f_t(x_i)= y_i)p(f_t(x_i))}{2},
\]
where
$f_t(x_i)$ is the predicted label of $x_i$.
The function $\mathrm{is}(\cdot)$ returns $+1$ if $\cdot$ is true and returns $-1$ otherwise, and $p$ denotes the probability (i.e., softmax output) for the predicted label $f_t(x_i)$.
Since $(\mathrm{is}(f_t(x_i)= y_i)p(f_t(x_i)))$ is in $[-1, 1]$, $\theta_{i,t}$ is in $[0,1]$.
\par
We experimentally verify 
that the total selection risk with $\btheta_1,\ldots,\btheta_t$ given by the above is smaller when the selected samples are cleaner. In other words, we verify that if we achieve a small (total) selection risk, the selected samples should be (averagely) cleaner.
We use MNIST training samples with 50\% label-noise and consider $0.5n$-set sample selection.
In this experiment, we assume that we know which samples are incorrectly labeled. 
We consider five subsets of samples, each of which contains $\{20\%, 40\%, 60\%, 80\%, 100\%\}$ of clean samples. For simplicity, we fixed the IDs of selected samples and trained DNN with the samples on each rate (i.e., $\bd_1,\ldots, \bd_T$ are the same at every epochs).
Then, we compared the total risks corresponding to the subsets of samples.
Fig.~\ref{fig:append_total_risk} shows the curves of the total selection risks along with the epochs, with $k=0.5n$ on MNIST. 
We can see that the subsets of the selected samples containing smaller incorrectly labeled samples had smaller total selection risks. In particular, 100\% clean subset of the samples achieved extremely low total selection risk compared to the others. Therefore, it is expected that we can train DNN with cleaner samples by aiming for smaller total selection risk.

\section{Result details}
\subsection{Effect of $\eta$ and $k$}
Table~\ref{table:hyper} shows the tuned hyperparameters $\eta$ and $k$. We can observe that the tuned $\eta$ was never the smallest of the range for which we searched. It suggests that the perturbation $\boldsymbol{r}$ plays an important role in finding cleaner samples. 
Moreover, it can be seen that the higher $\eta$ was used for noisier cases of CIFAR-10 and CIFAR-100. It suggests that \Ours needs to have a larger perturbation to explore (minor) clean samples in cases with higher noise rates.
\par
Table~\ref{table:hyper} also shows that the tuned $k$s were always less than or equal to the ideal $k$ (that corresponds to the true noise rate) in all cases. This indicates that it is better to select a limited number of certainly clean samples while avoiding contamination by incorrectly labeled samples.

\subsection{Test accuracy and label precision curves}
\begin{figure}[t]
\centering
\includegraphics[width=1\linewidth]{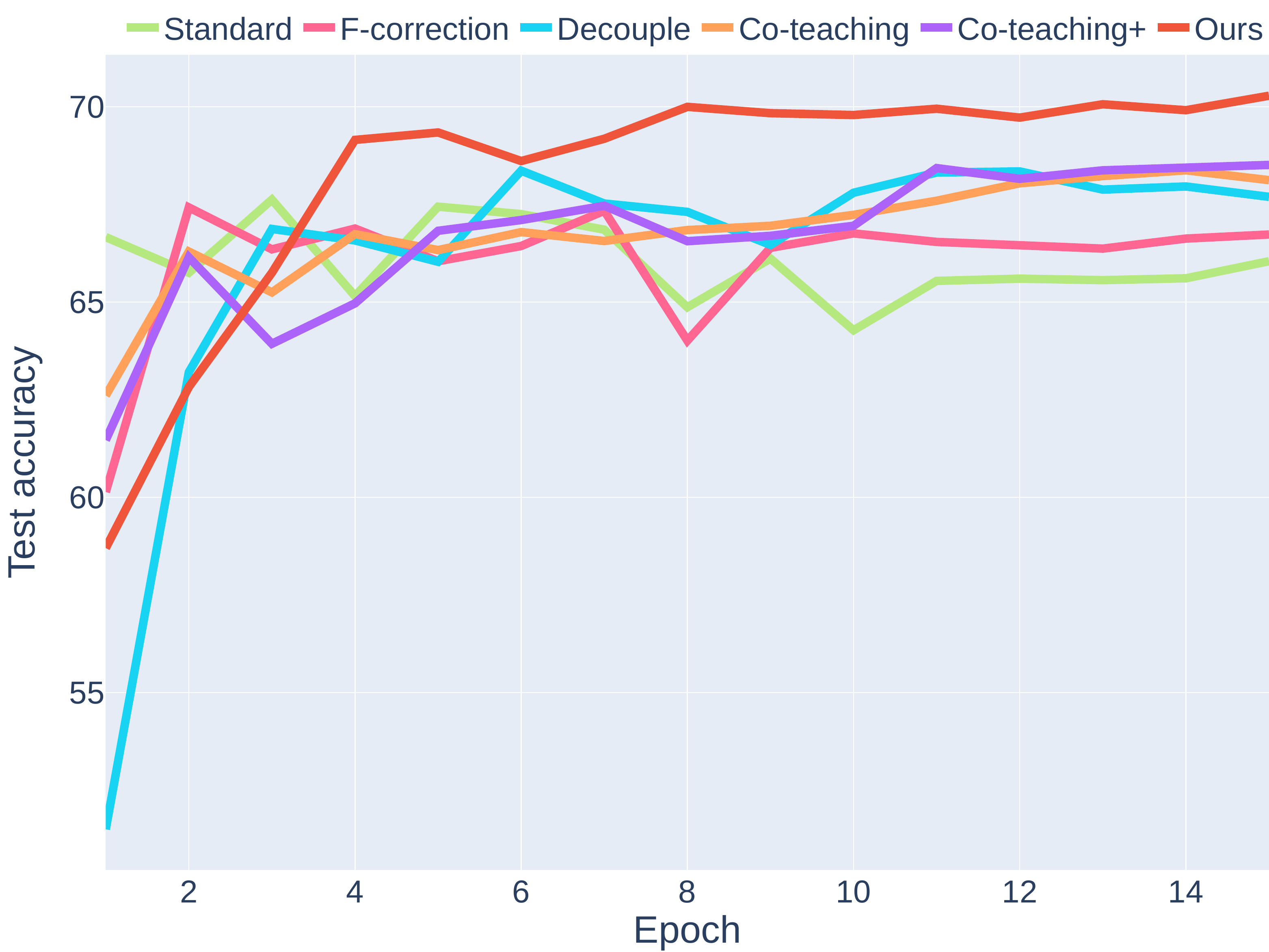}
\caption{Test accuracy curves on Clothing1M.}
\label{fig:append_clothing_graph}
\end{figure}
\begin{figure*}[t]
\centering
\includegraphics[width=1\linewidth]{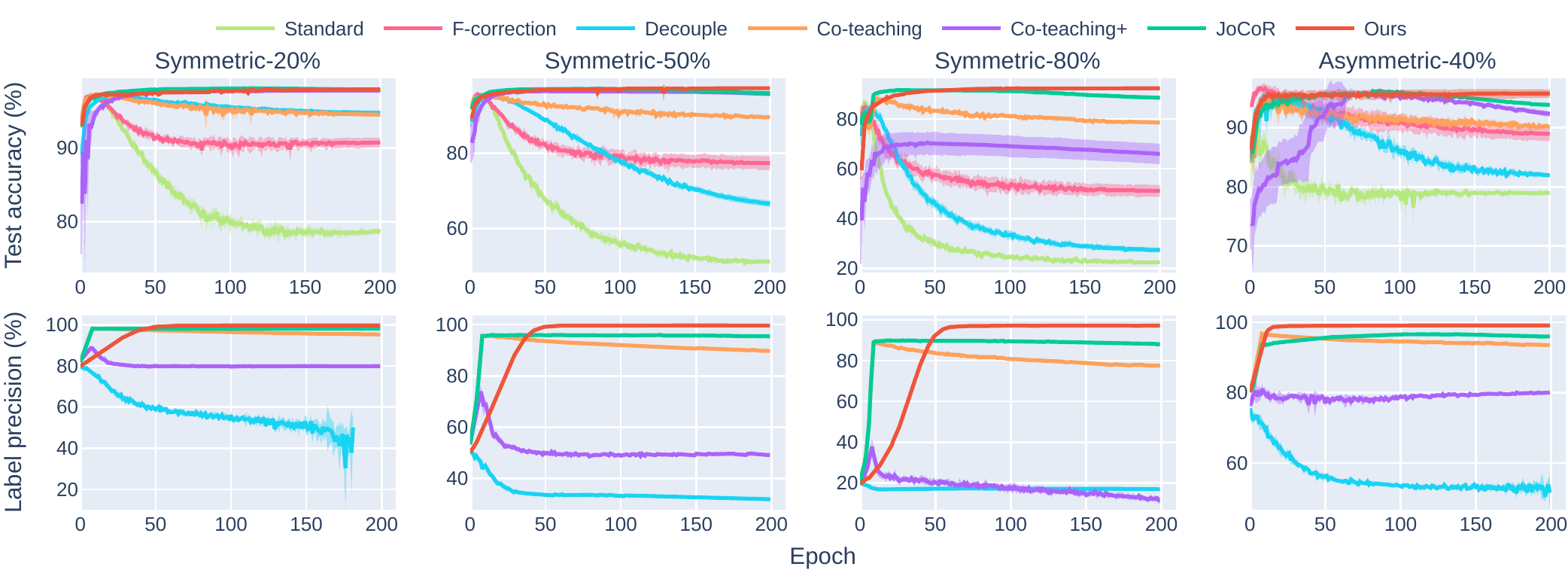}
\caption{Test accuracy (Top) and label precision (Bottom) curves with error band on MNIST.}
\label{fig:append_mnist_graph}
\end{figure*}
\begin{figure*}[t]
\centering
\includegraphics[width=1\linewidth]{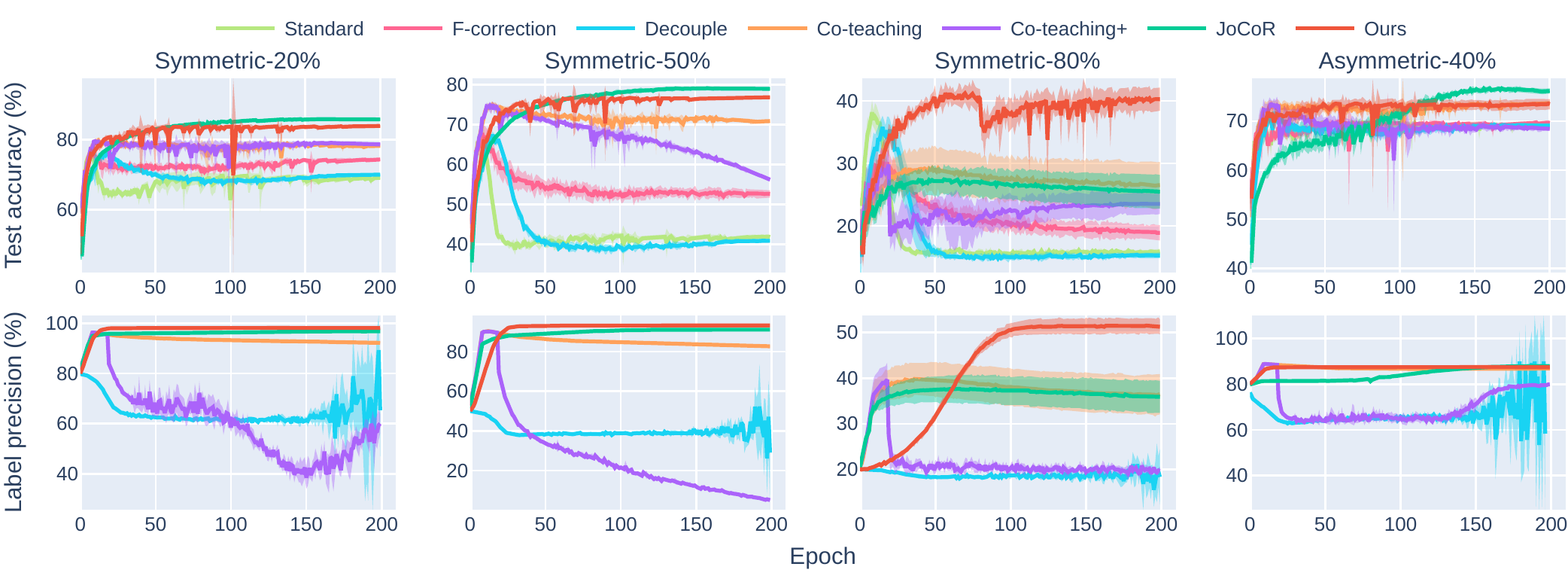}
\caption{Test accuracy (Top) and label precision (Bottom) curves with error band on CIFAR-10.}
\label{fig:append_cifar10_graph}
\end{figure*}
\begin{figure*}[t]
\centering
\includegraphics[width=1\linewidth]{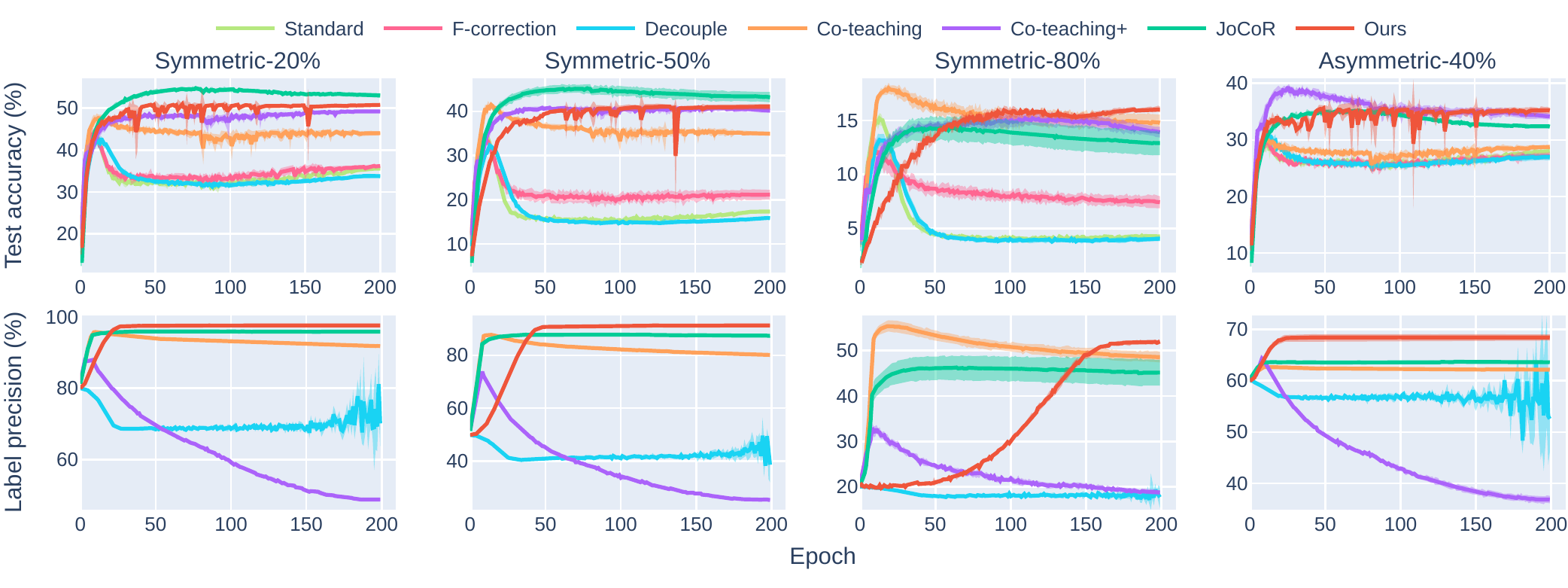}
\caption{Test accuracy (Top) and label precision (Bottom) curves with error band on CIFAR-100.}
\label{fig:append_cifar100_graph}
\end{figure*}

Figs.~\ref{fig:append_mnist_graph},~\ref{fig:append_cifar10_graph}, and~\ref{fig:append_cifar100_graph} show test accuracy and label precision curves on MNIST, CIFAR-10, and CIFAR-100, respectively.
On most datasets, the test accuracy and label precision of \Ours kept increasing or saturating as the DNN was trained. On the other hand, in the Symmetric-80\% case on MNIST and CIFAR-10, or in most cases on CIFAR-100, both test accuracy and label precision of baselines gradually decreased after they reached their peak. \par

Fig.~\ref{fig:append_clothing_graph} shows the test accuracy curves on Clothing1M. Note that, for Clothing1M, the results of JoCoR are not plotted because we could not reproduce the performance presented in their original paper due to the lack of hyperparameter value information, although we followed that JoCoR is trained for Clothing1M just by 15 epochs. Similar to other datasets, \Ours achieved the best test accuracy compared to the baselines. In addition, from getting the best performance at the end of the training, i.e., at the 15th epoch, we can expect to get better performance by training DNN with \Ours for a longer epoch.

\subsection{Change of noise-risk}
\begin{figure*}[t]
\centering
\begin{subfigure}{.24\textwidth}
        \centering
        Symmetric-20\%
        \includegraphics[width=\linewidth]{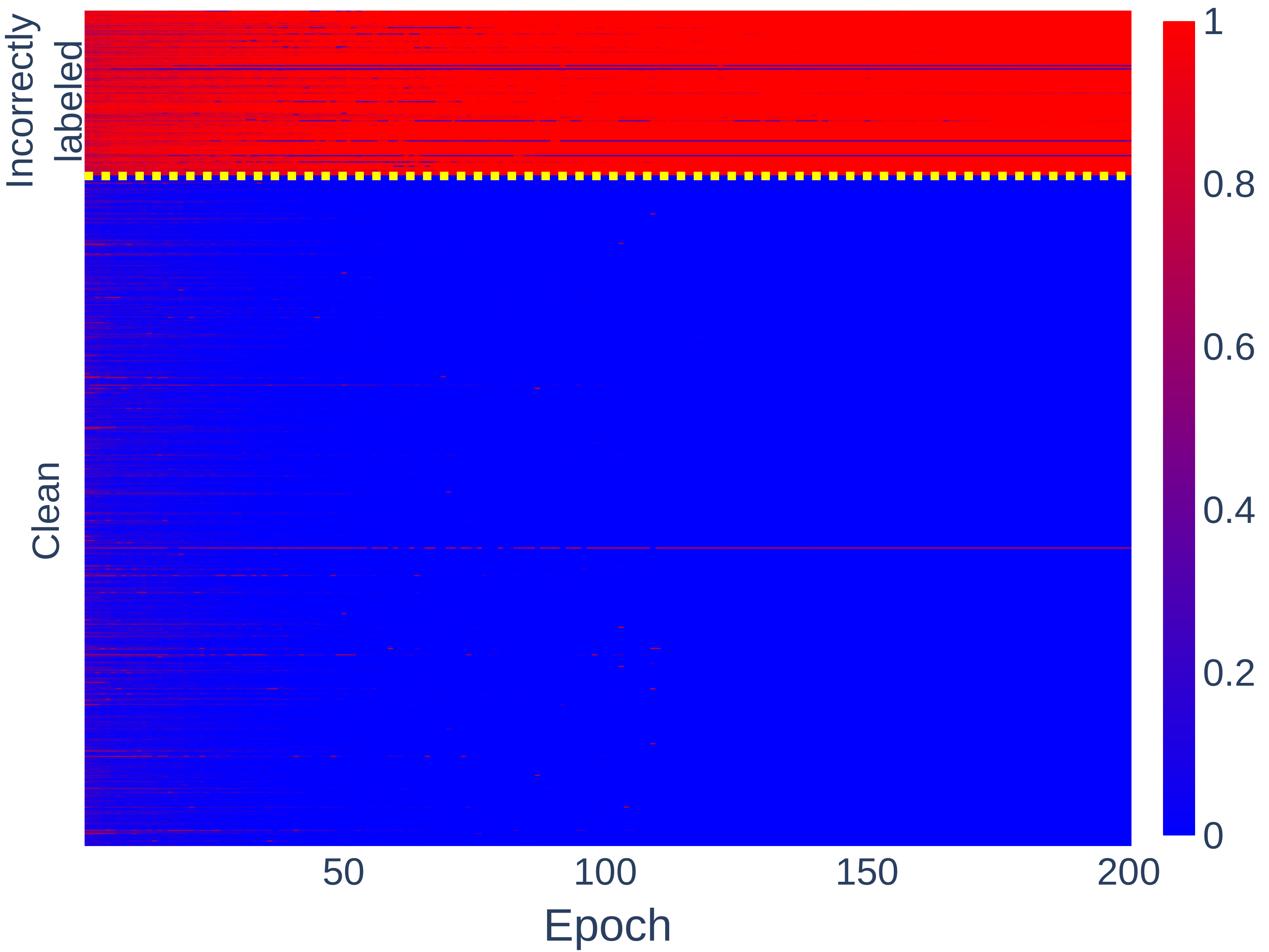}
\end{subfigure}\hfill
\begin{subfigure}{.24\textwidth}
        \centering
        Symmetric-50\%
        \includegraphics[width=\linewidth]{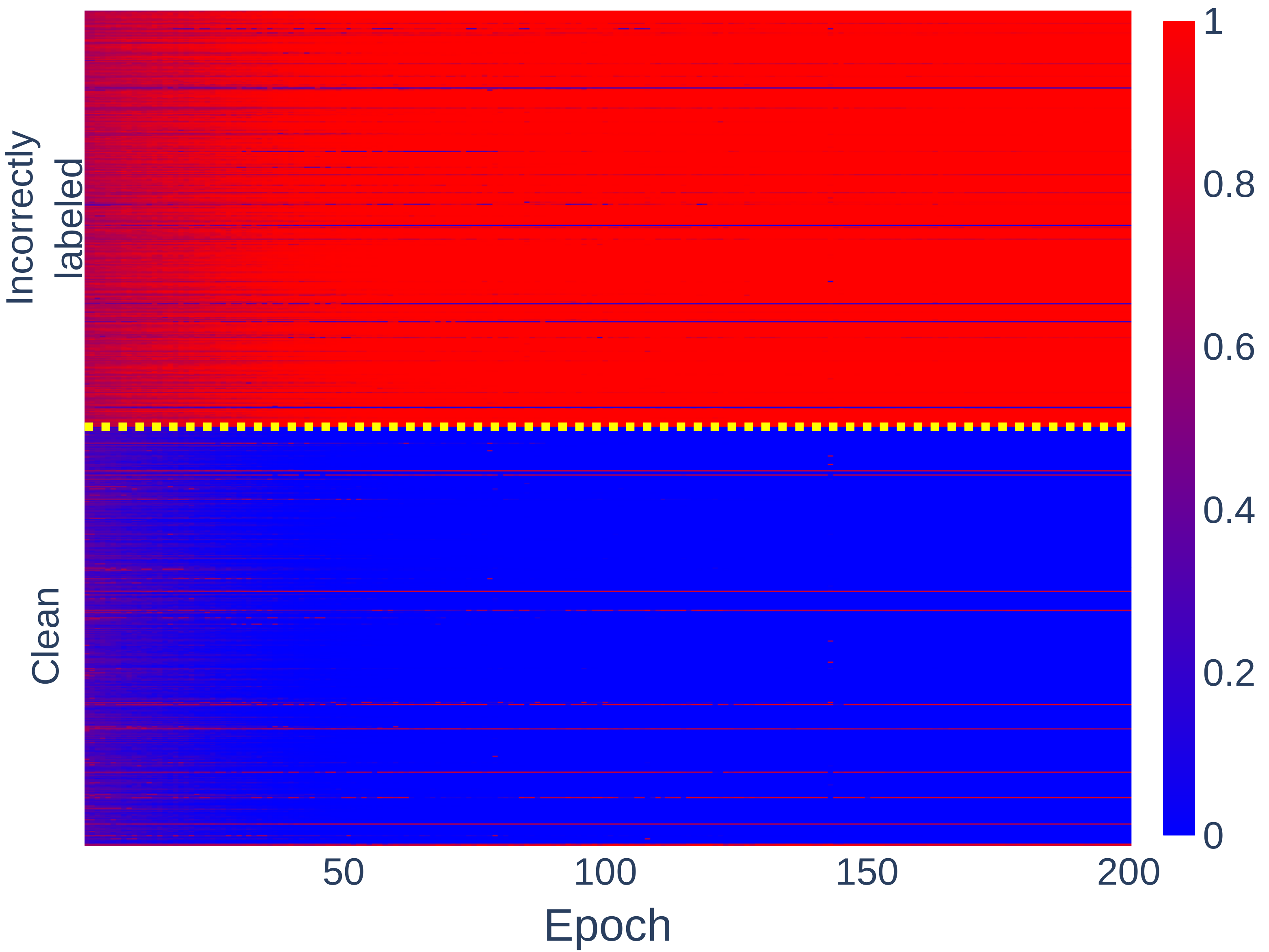}
\end{subfigure}\hfill
\begin{subfigure}{.24\textwidth}
        \centering
        Symmetric-80\%
        \includegraphics[width=\linewidth]{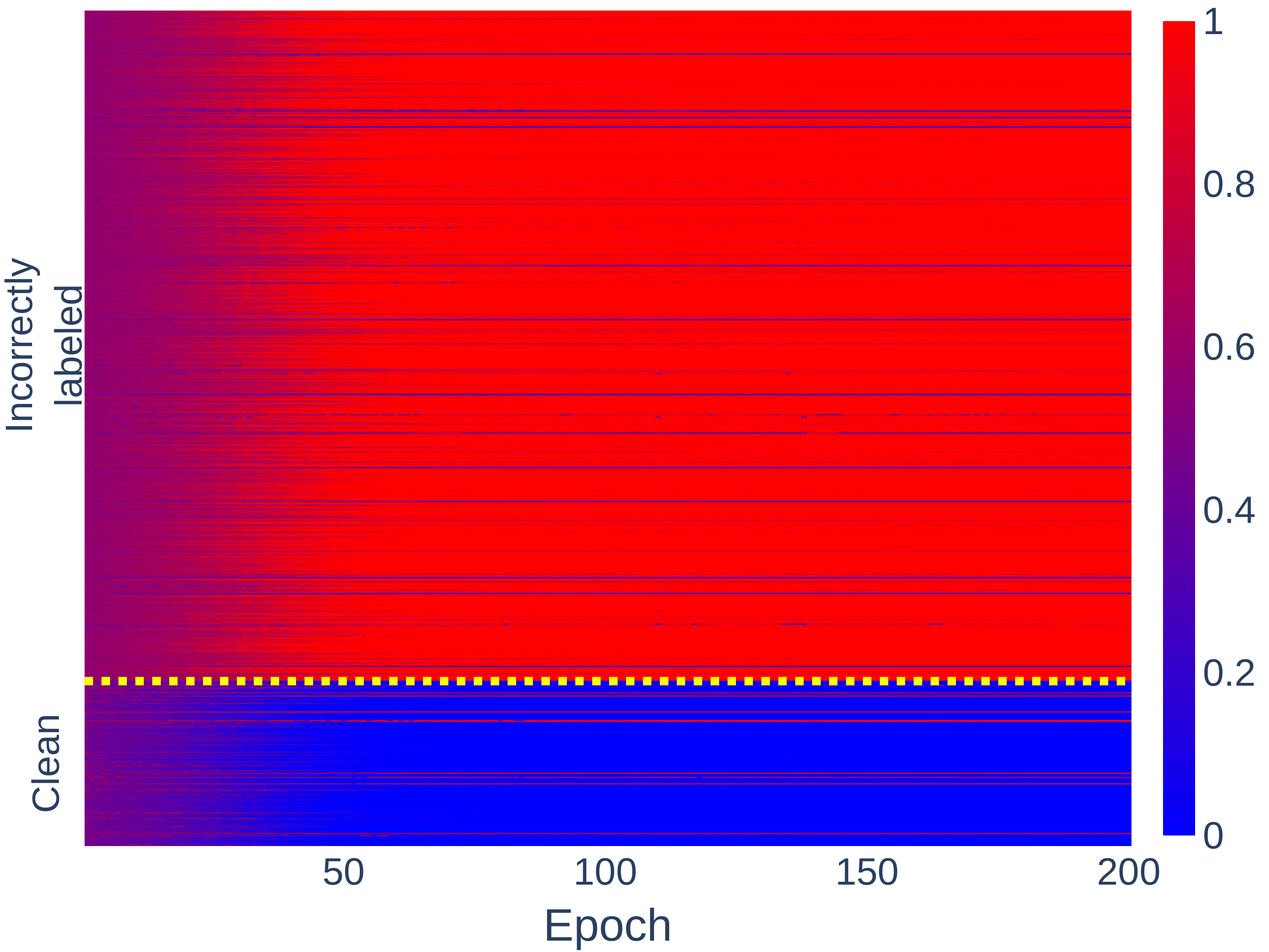}
\end{subfigure}\hfill
\begin{subfigure}{.24\textwidth}
        \centering
        Asymmetric-40\%
        \includegraphics[width=\linewidth]{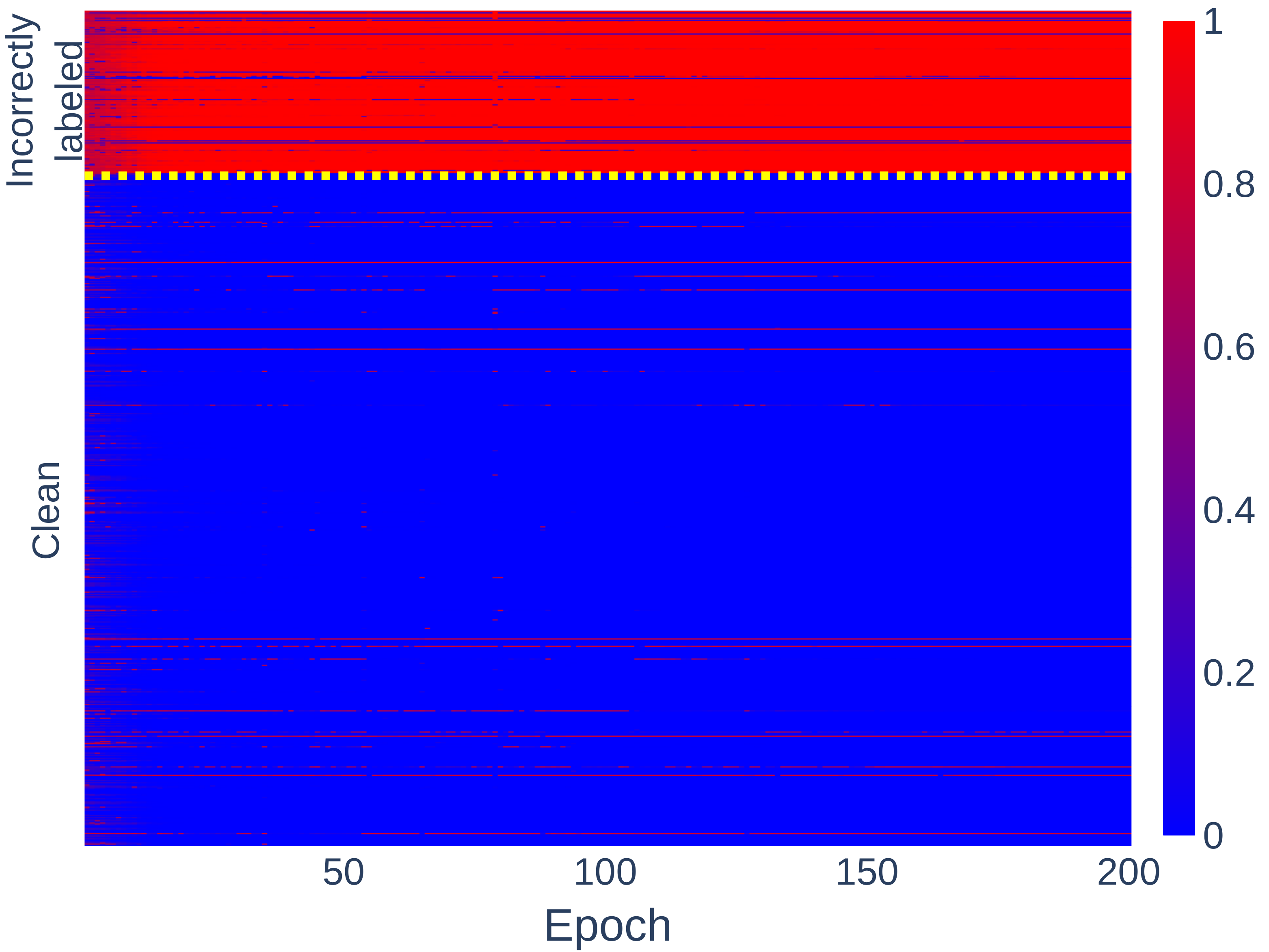}
\end{subfigure}\hfill
\caption{Estimated noise-risk on MNIST.}
\label{fig:append_mnist_loss}
\end{figure*}

\begin{figure*}[t]
\centering
\begin{subfigure}{0.24\textwidth}
        \centering
        Symmetric-20\%
        \includegraphics[width=\linewidth]{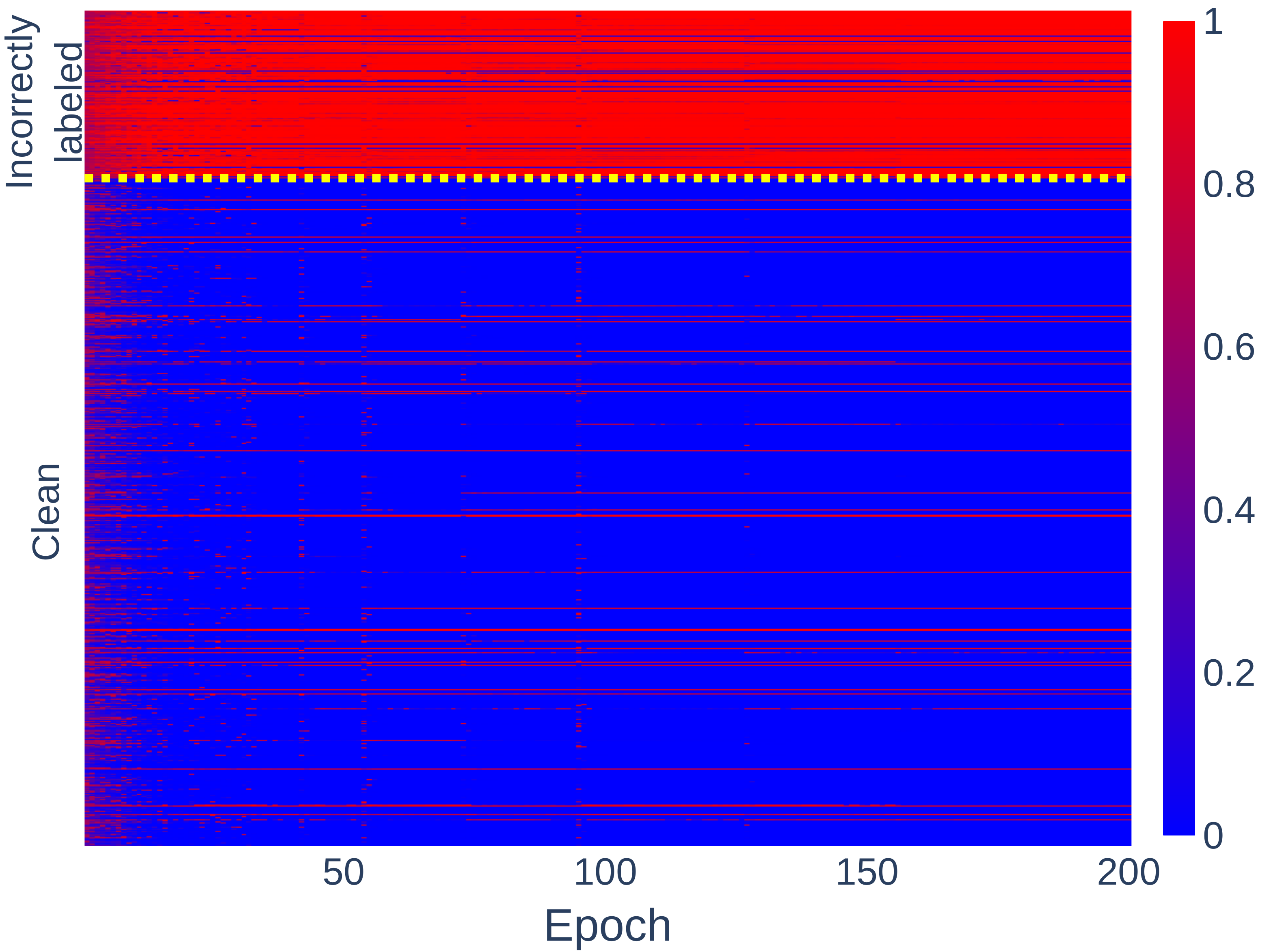}
\end{subfigure}\hfill
\begin{subfigure}{0.24\textwidth}
        \centering
        Symmetric-50\%
        \includegraphics[width=\linewidth]{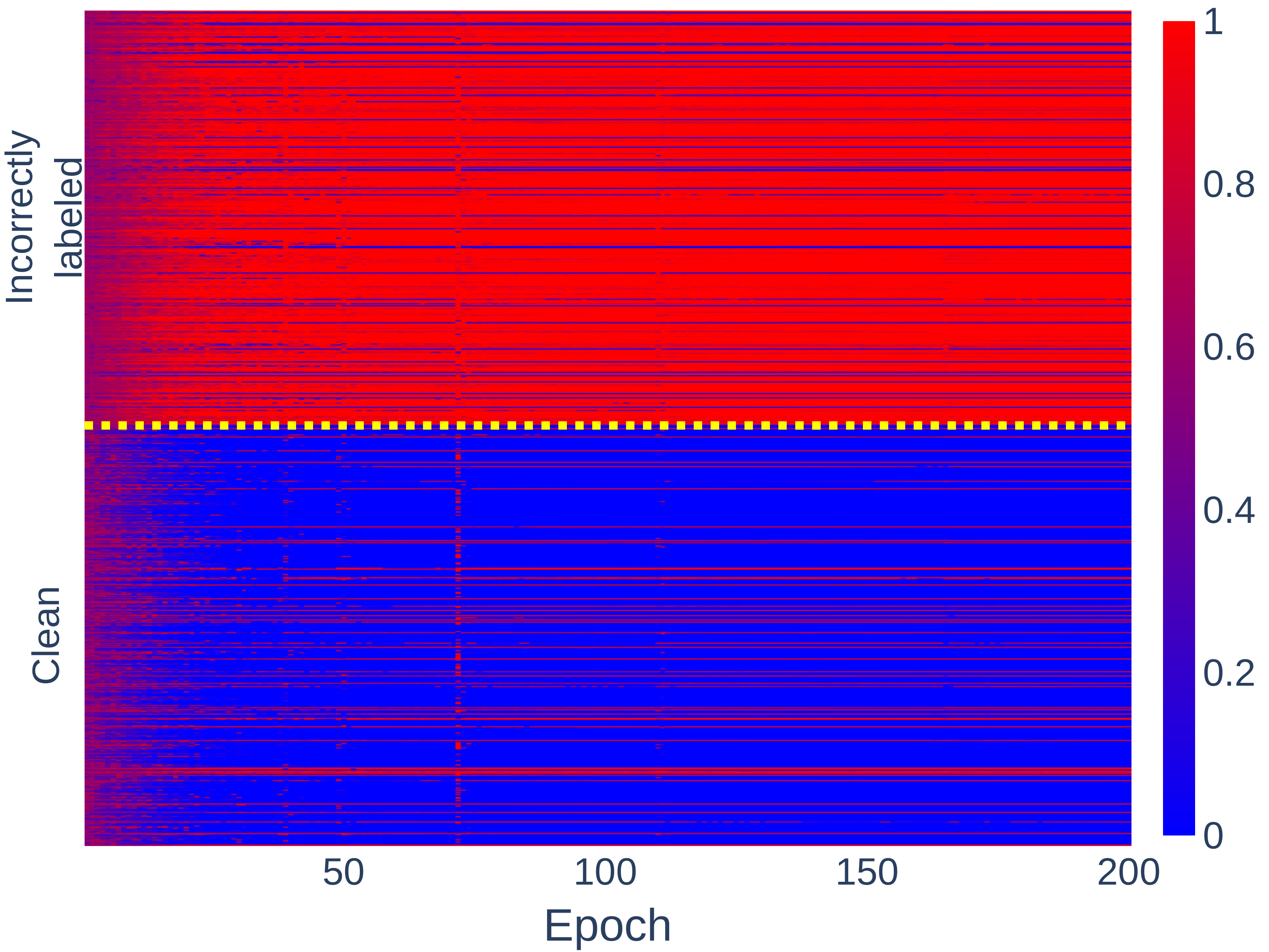}
\end{subfigure}\hfill
\begin{subfigure}{0.24\textwidth}
        \centering
        Symmetric-80\%
        \includegraphics[width=\linewidth]{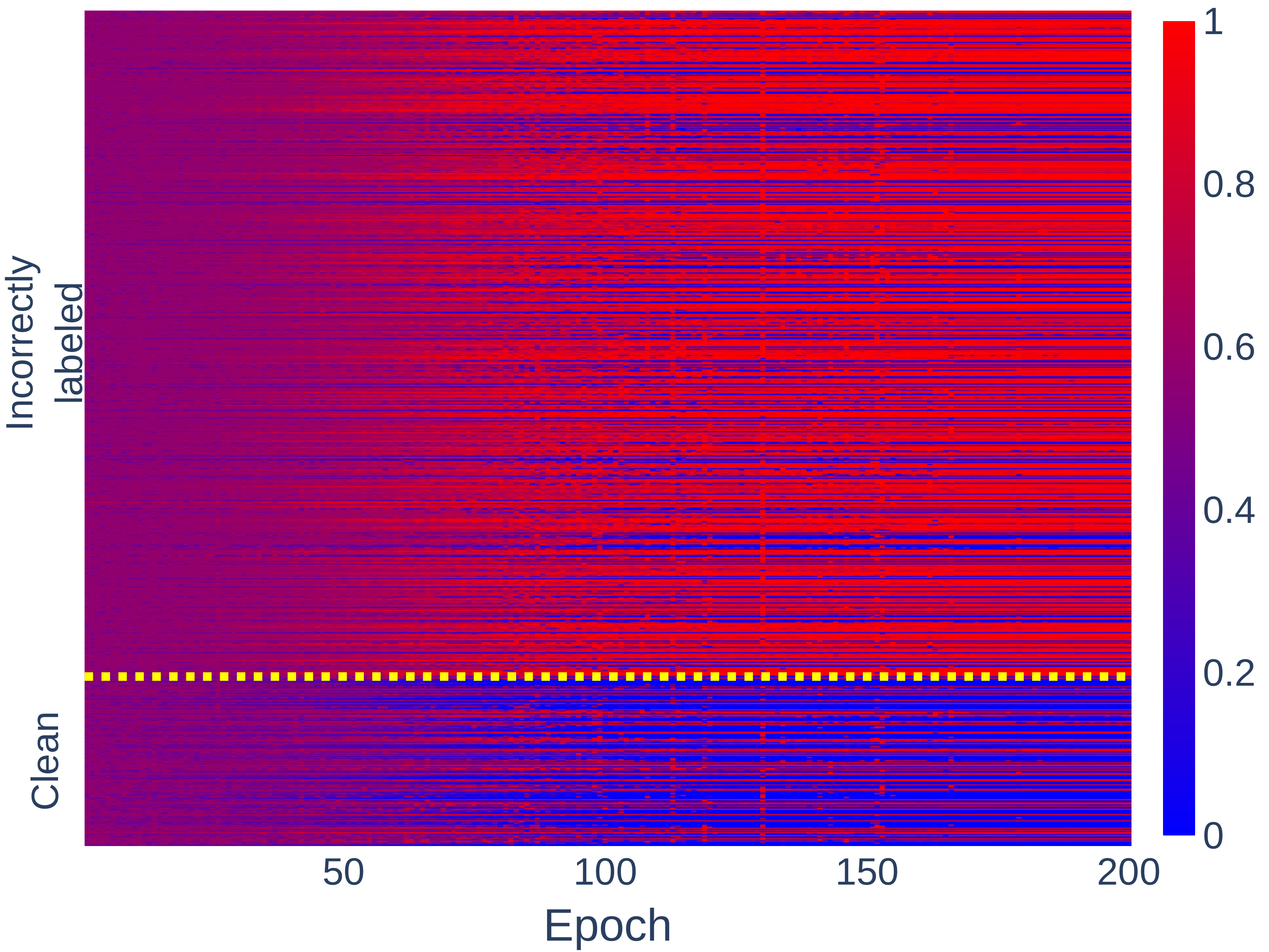}
\end{subfigure}\hfill
\begin{subfigure}{0.24\textwidth}
        \centering
        Asymmetric-40\%
        \includegraphics[width=\linewidth]{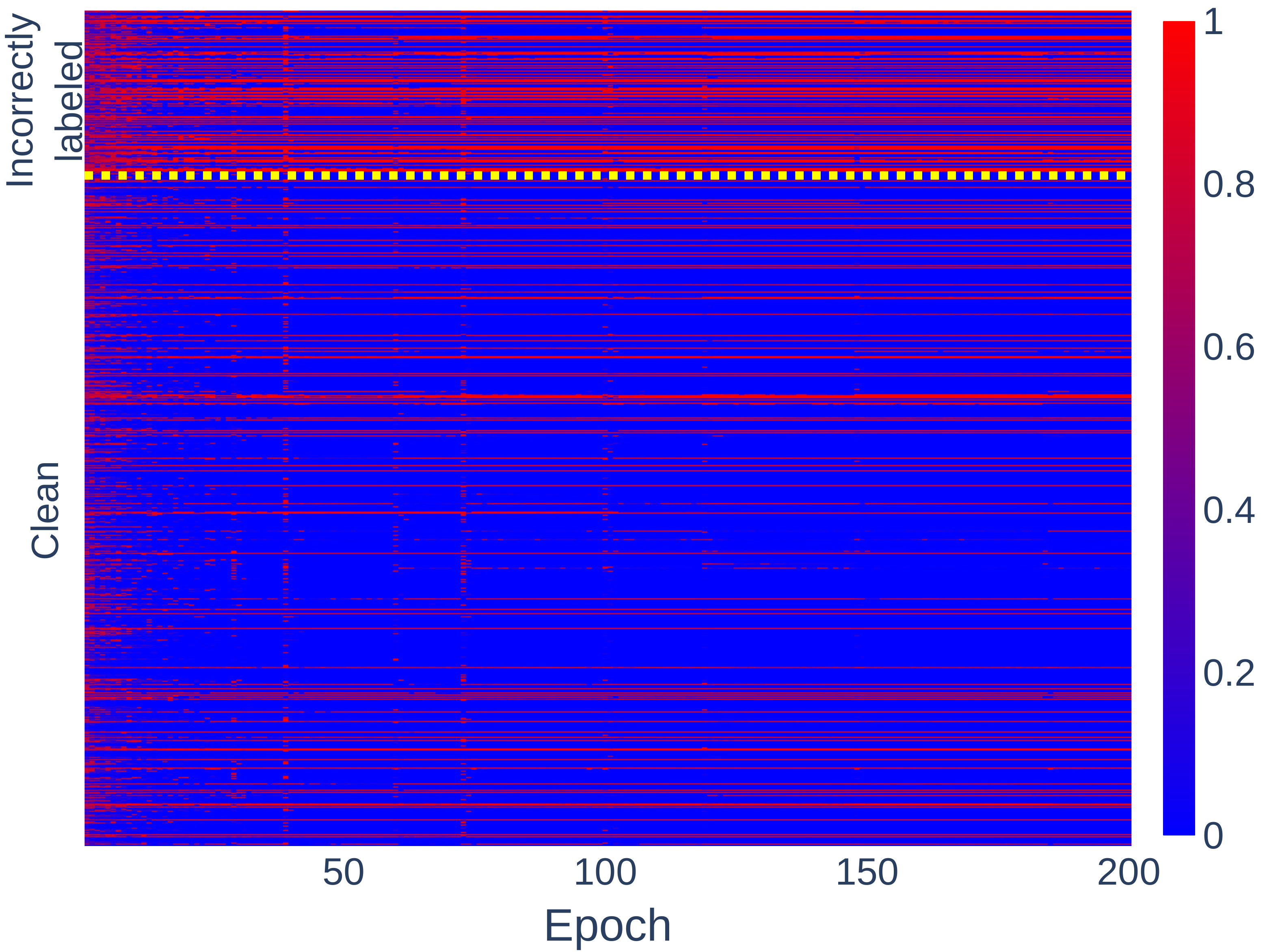}
\end{subfigure}\hfill
\caption{Estimated noise-risk on CIFAR-10.}
\label{fig:append_cifar10_loss}
\end{figure*}

\begin{figure*}[t]
\centering
\begin{subfigure}{0.24\textwidth}
        \centering
        Symmetric-20\%
        \includegraphics[width=\linewidth]{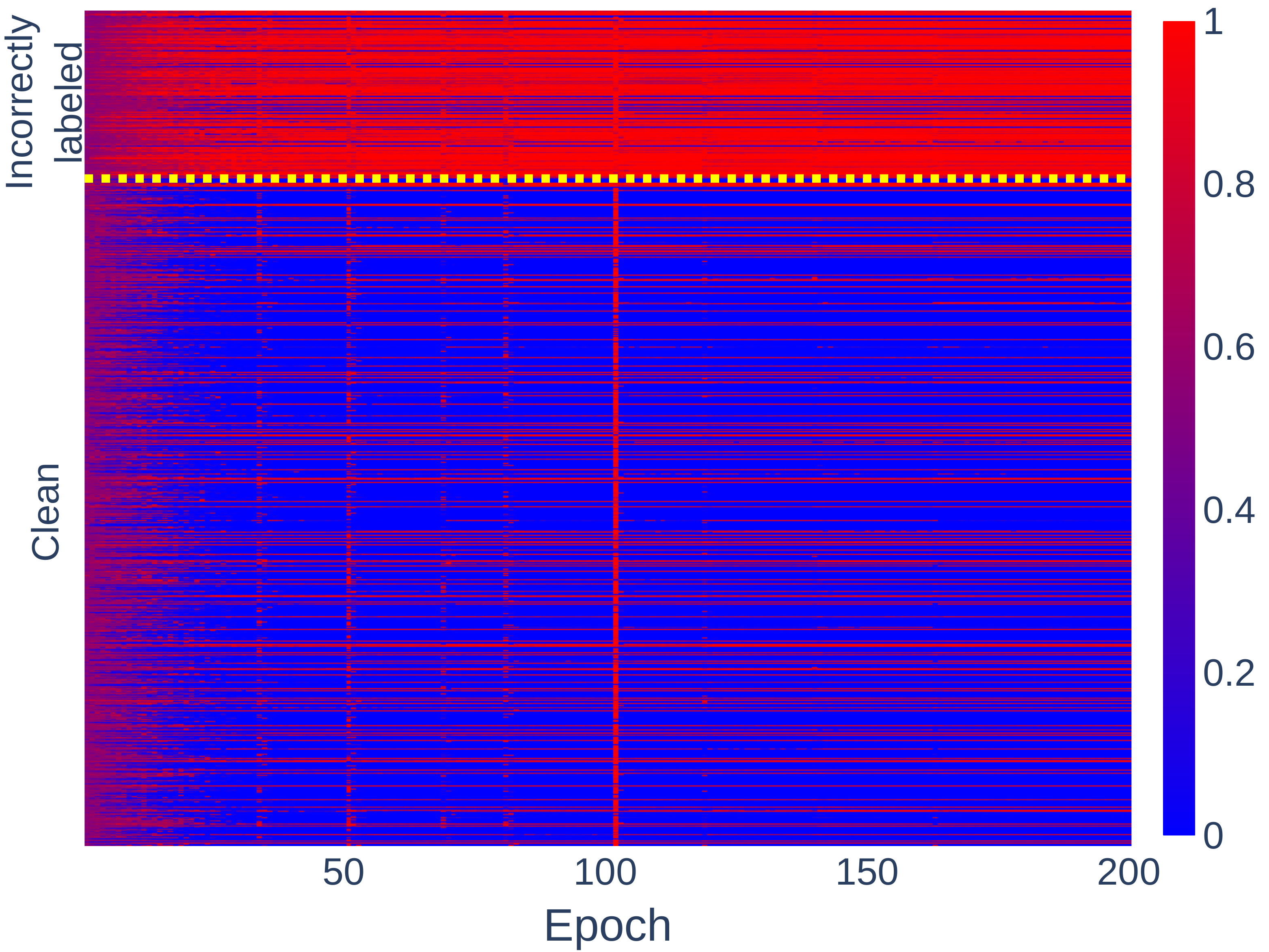}
\end{subfigure}\hfill
\begin{subfigure}{0.24\textwidth}
        \centering
        Symmetric-50\%
        \includegraphics[width=\linewidth]{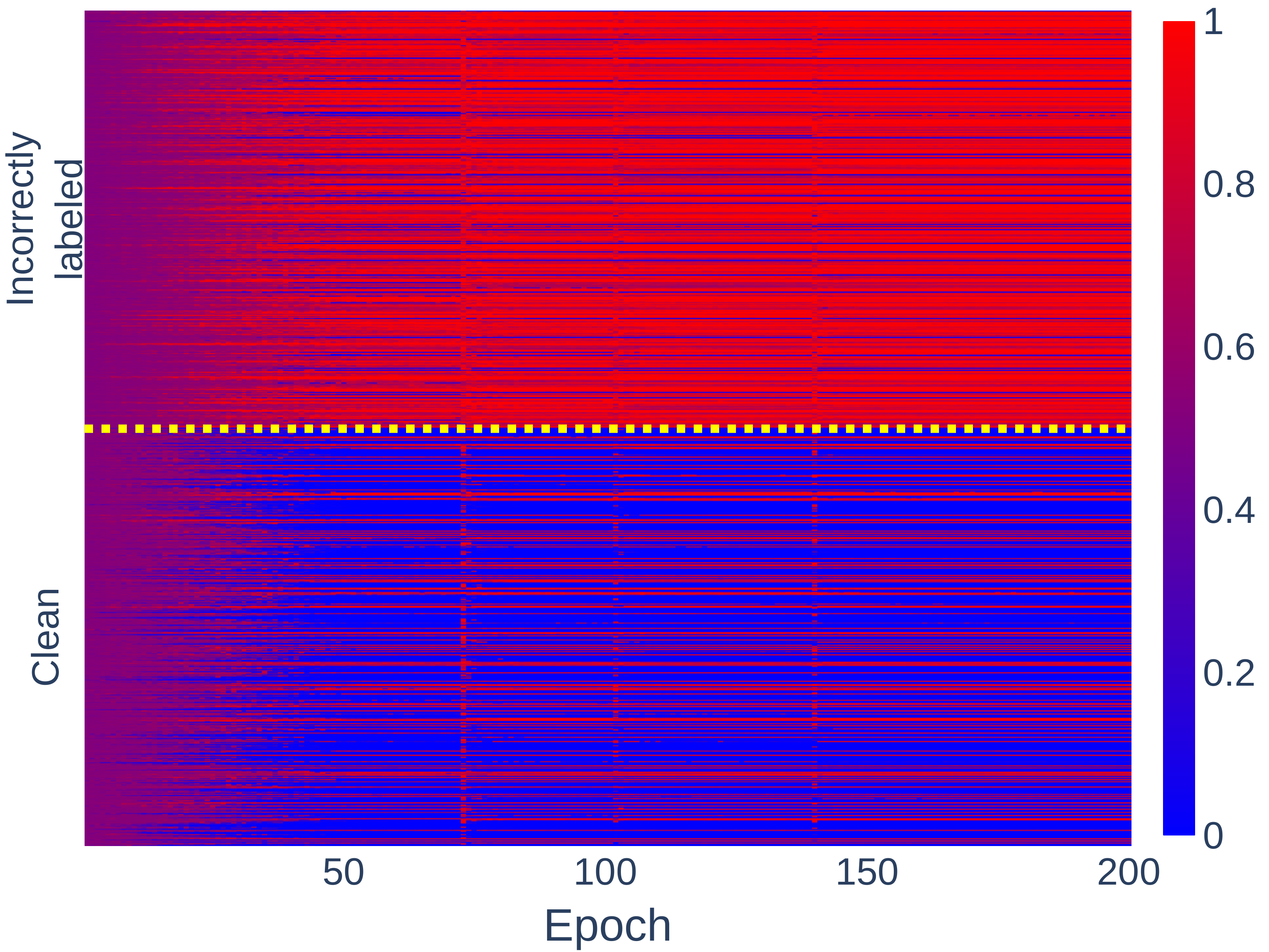}
\end{subfigure}\hfill
\begin{subfigure}{0.24\textwidth}
        \centering
        Symmetric-80\%
        \includegraphics[width=\linewidth]{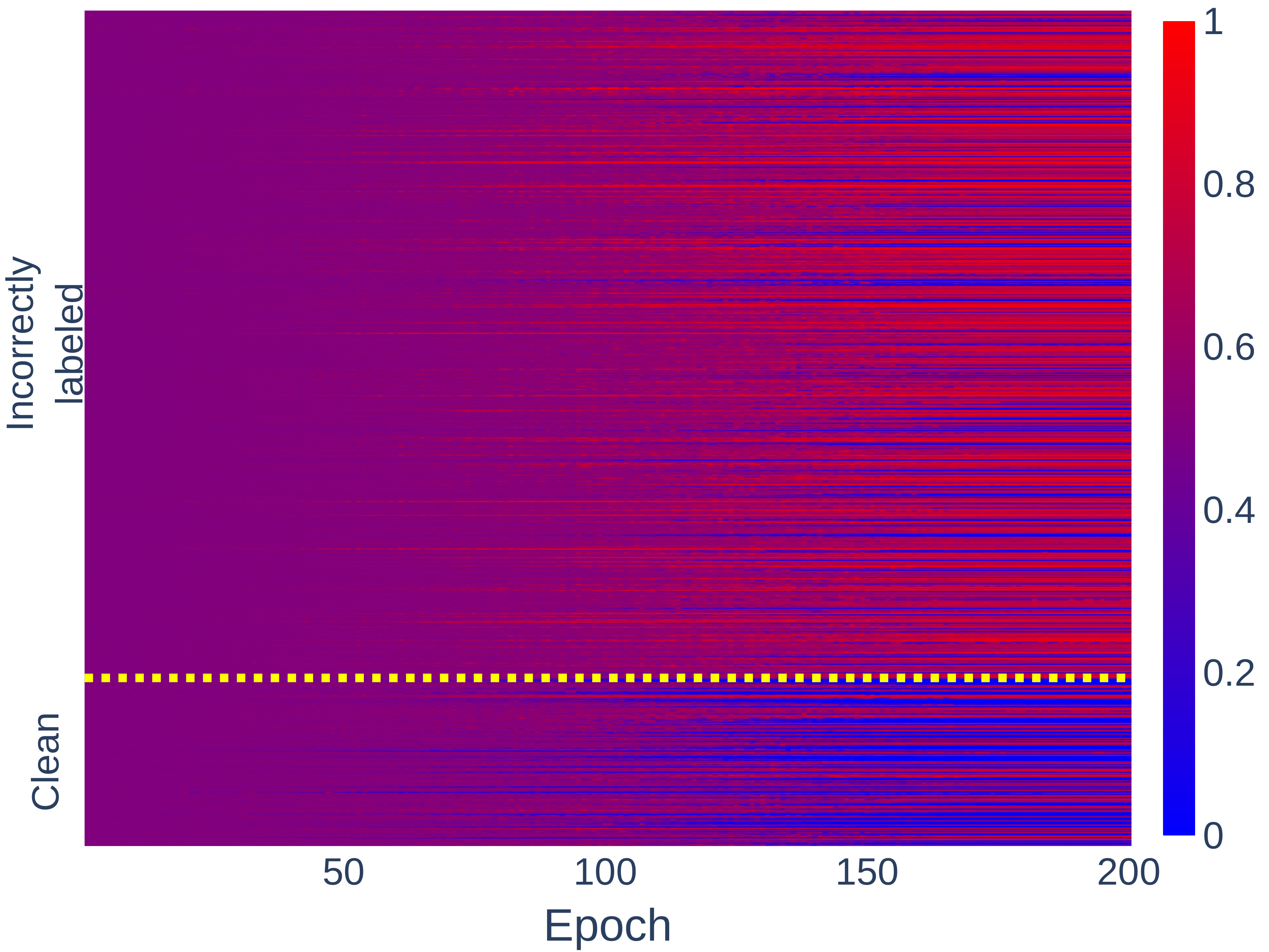}
\end{subfigure}\hfill
\begin{subfigure}{0.24\textwidth}
        \centering
        Asymmetric-40\%
        \includegraphics[width=\linewidth]{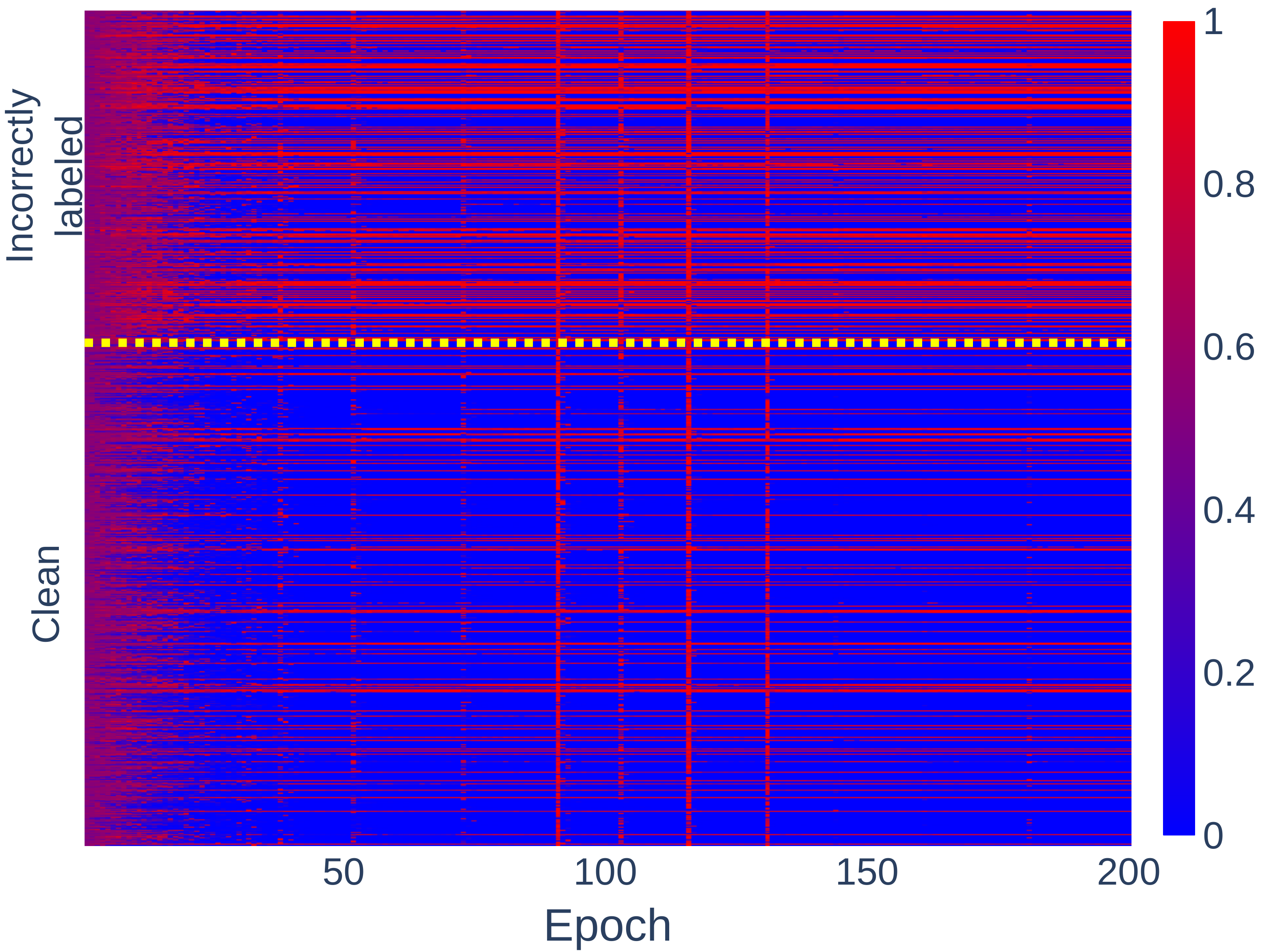}
\end{subfigure}\hfill
\caption{Estimated noise-risk on CIFAR-100.}
\label{fig:append_cifar100_loss}
\end{figure*}

Figs.~\ref{fig:append_mnist_loss},~\ref{fig:append_cifar10_loss}, and~\ref{fig:append_cifar100_loss} show the change of the noise-risk vector to each training sample along with epochs on MNIST, CIFAR-10, and CIFAR-100, respectively.
On MNIST, the noise-risks of clean samples and incorrectly labeled samples were distinguished quickly. On CIFAR-10 and CIFAR-100, as the training sample contained more incorrectly-labeled samples, it became slower to distinguish between clean samples and incorrectly labeled samples, and relatively many samples are incorrectly distinguished. In particular, in the Symmetric-80\% case on CIFAR-100, clean samples and incorrectly labeled samples were not well distinguished even up to 100 epochs. However, even in this case, the clean samples had smaller risk in the end, and \Ours could achieve higher precision than the baselines.

\subsection{Feature distributions}
\begin{figure*}[t]
\centering
\begin{subfigure}{.49\textwidth}
        \centering
        Symmetric-20\%
        \includegraphics[width=\linewidth]{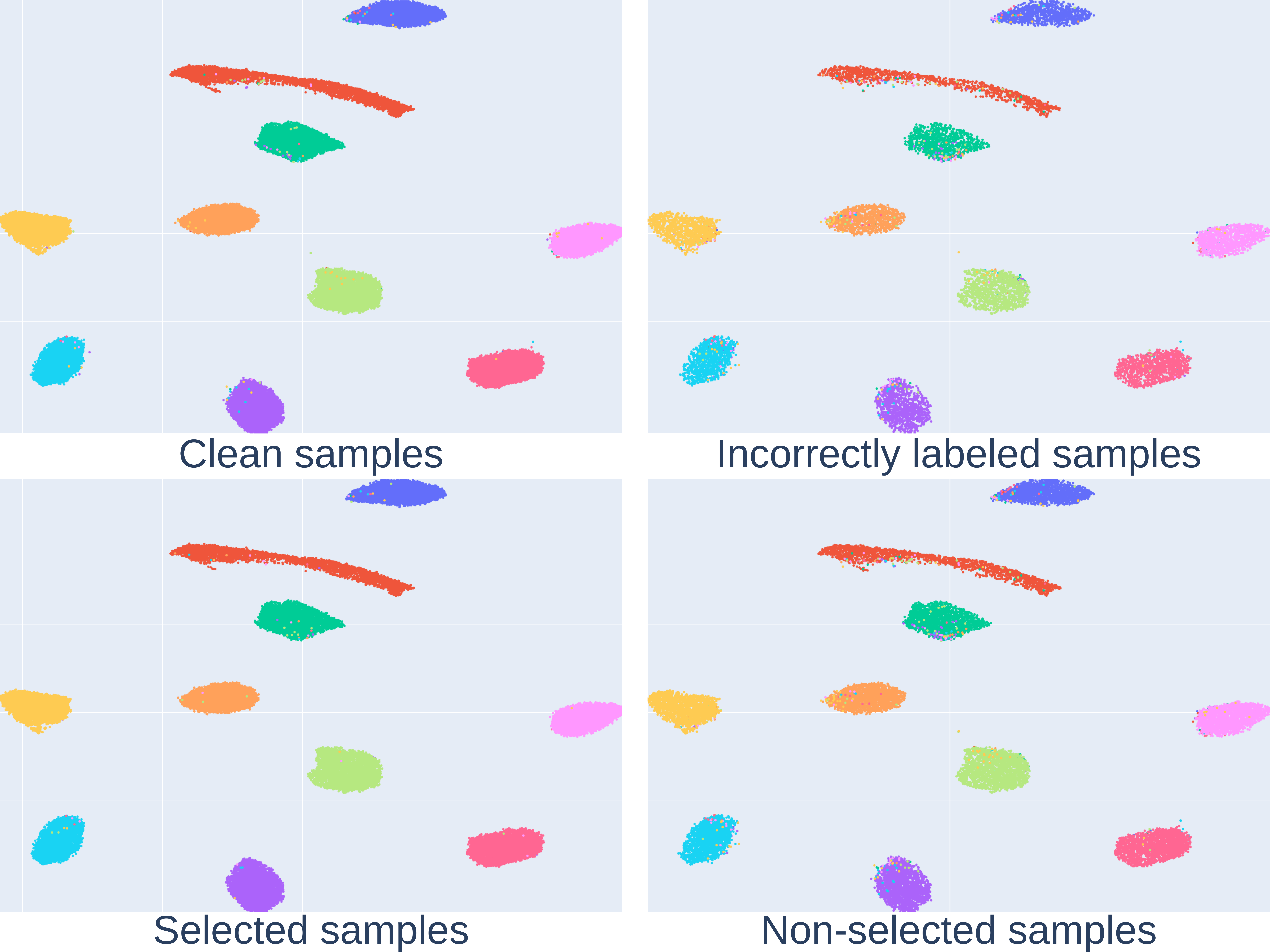}
\end{subfigure}\hfill
\begin{subfigure}{.49\textwidth}
        \centering
        Symmetric-50\%
        \includegraphics[width=\linewidth]{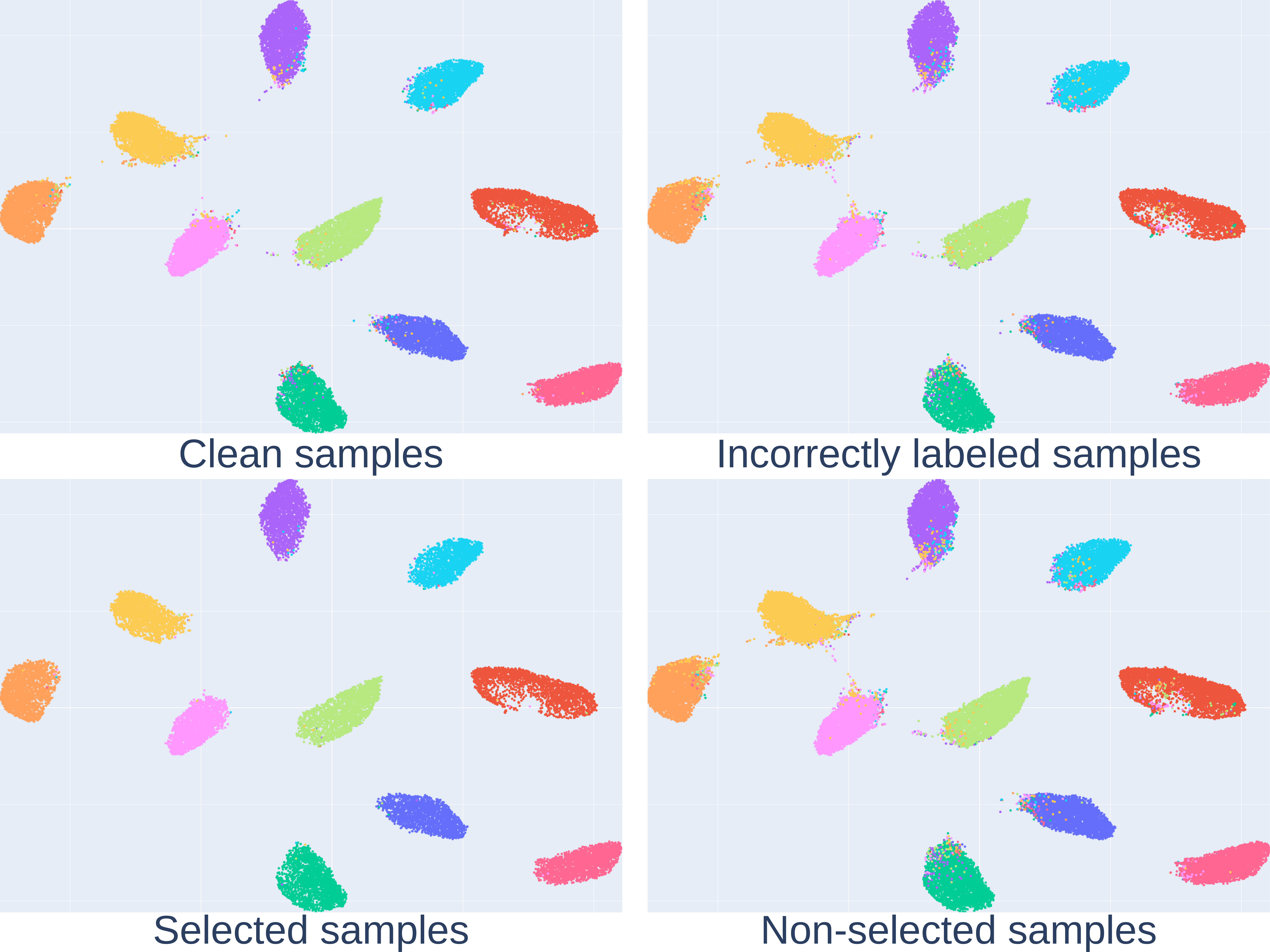}
\end{subfigure}\hfill
\par\bigskip
\begin{subfigure}{.49\textwidth}
        \centering
        Symmetric-80\%
        \includegraphics[width=\linewidth]{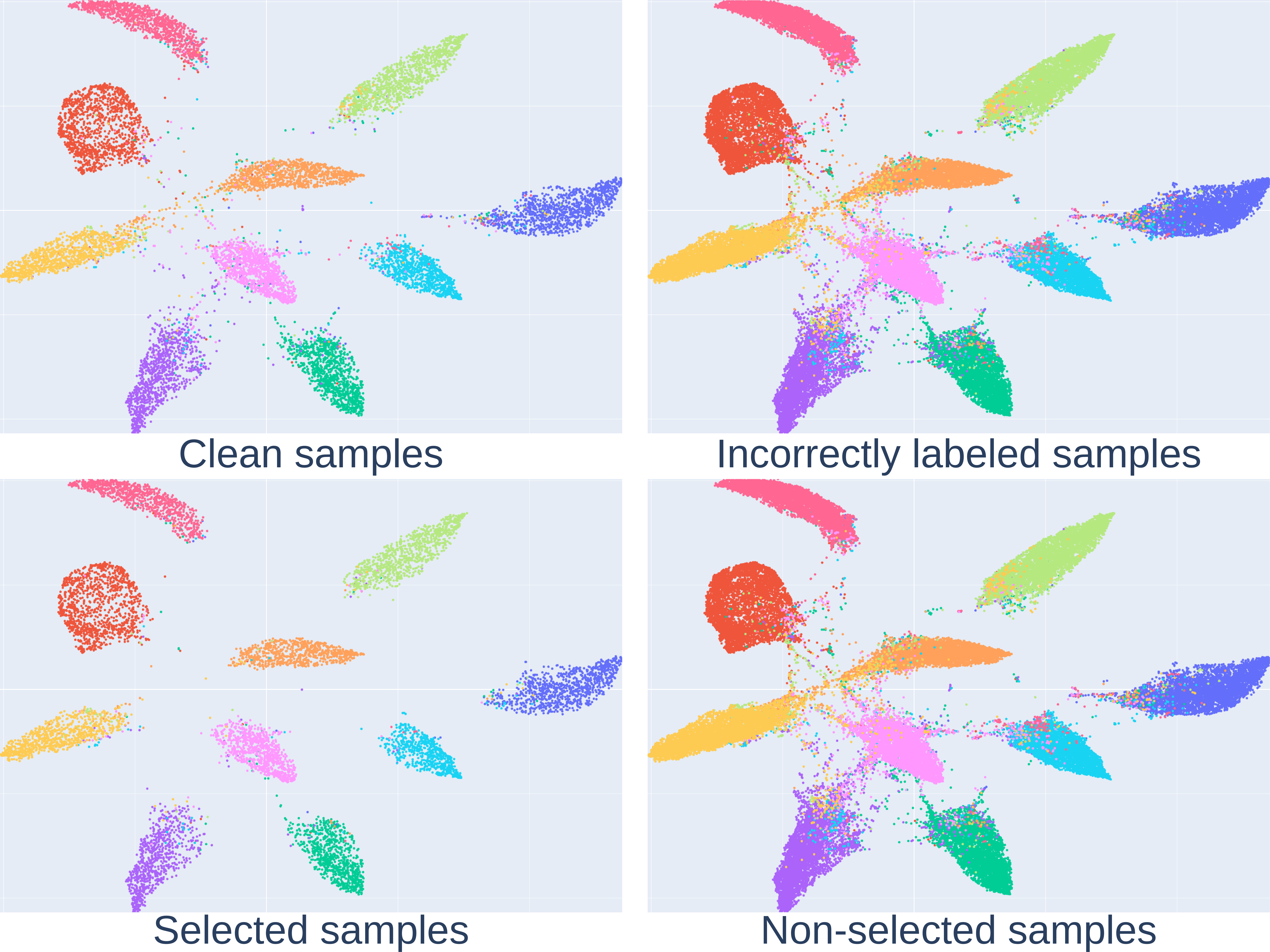}
\end{subfigure}\hfill
\begin{subfigure}{.49\textwidth}
        \centering
        Asymmetric-40\%
        \includegraphics[width=\linewidth]{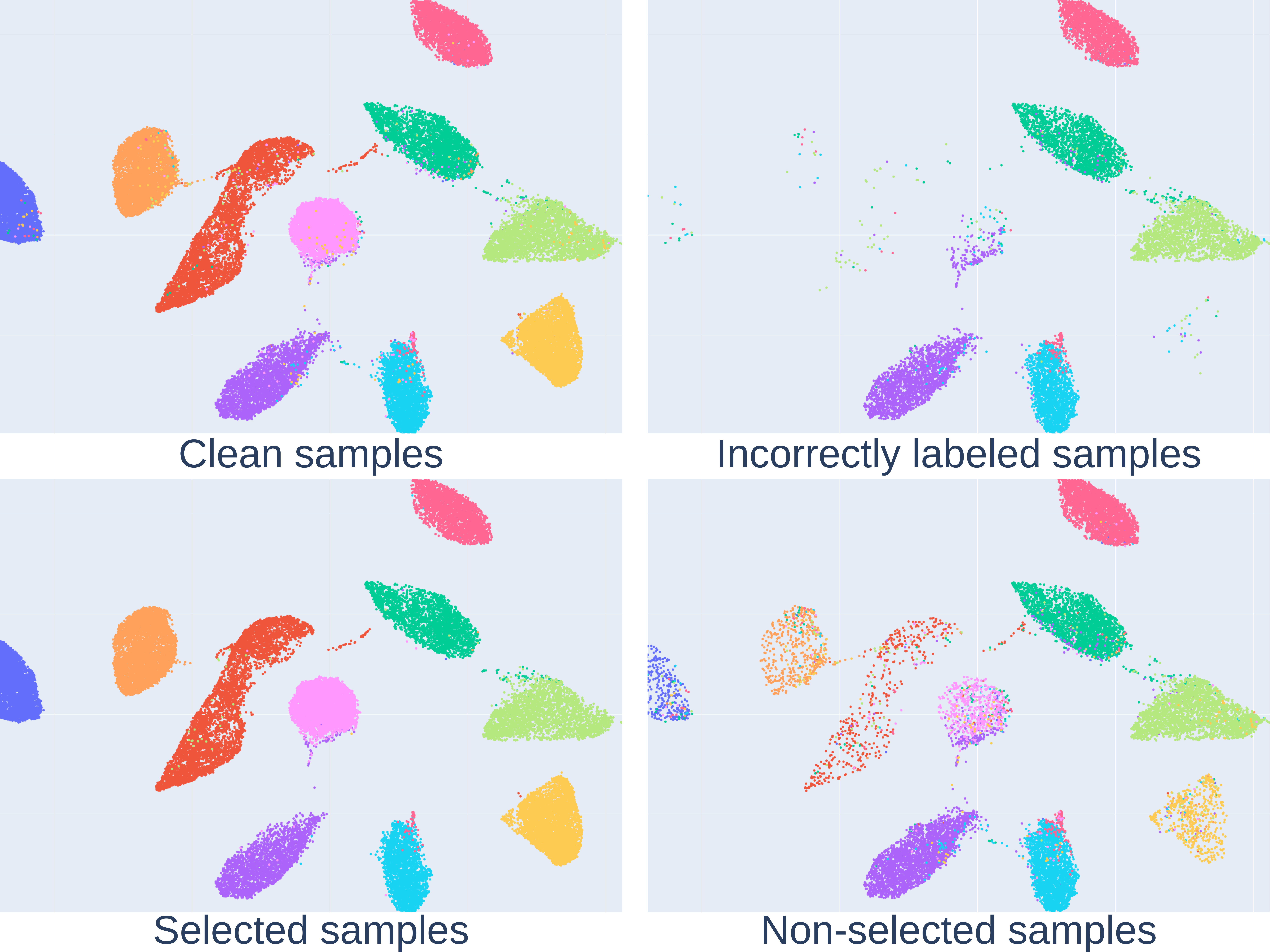}
\end{subfigure}\hfill
\caption{Visualization of the feature distribution of the noisy training samples on MNIST.}
\label{fig:append_mnist_samples}
\end{figure*}

\begin{figure*}[t]
\centering
\begin{subfigure}{.49\textwidth}
        \centering
        Symmetric-20\%
        \includegraphics[width=\linewidth]{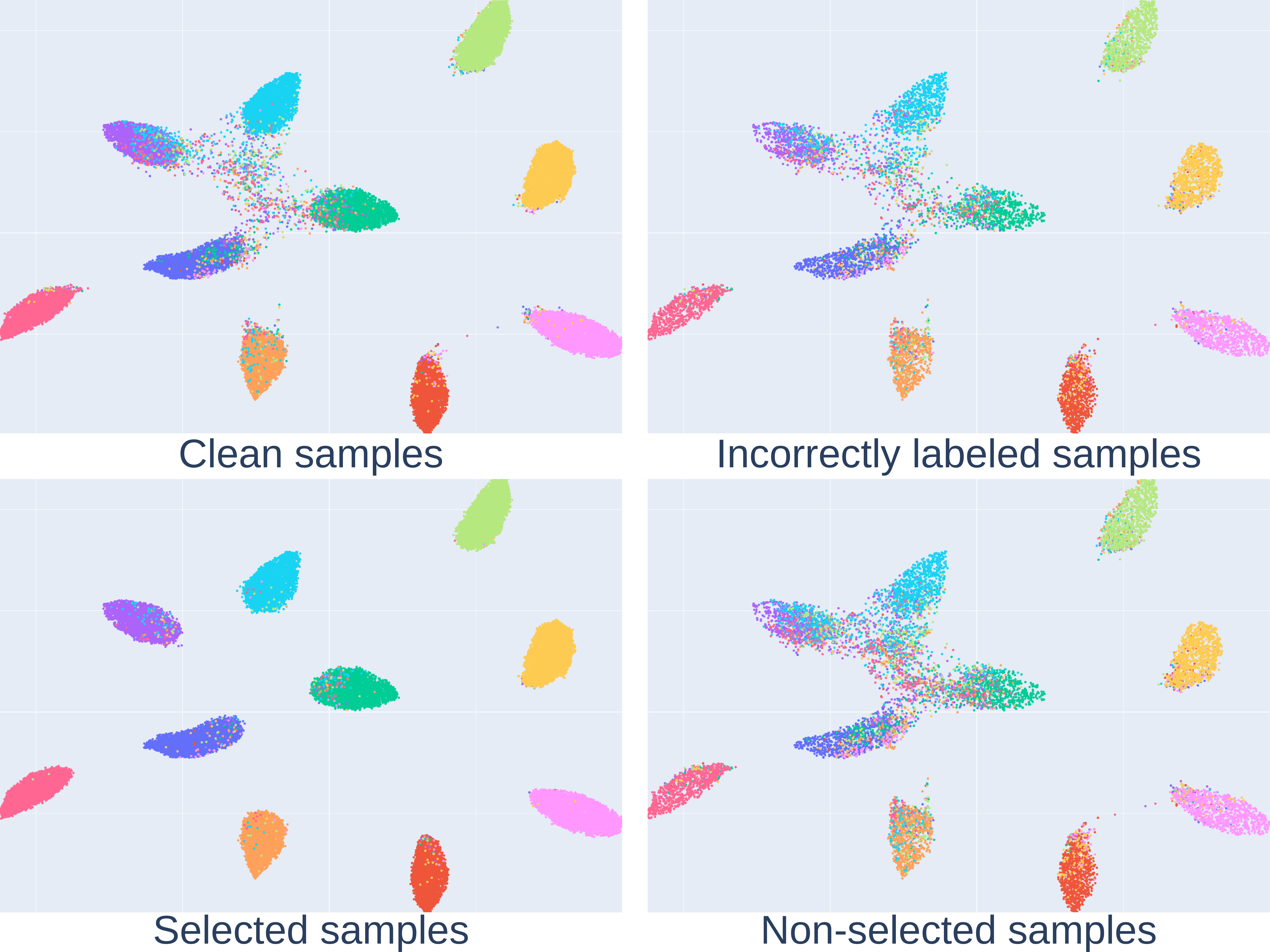}
\end{subfigure}\hfill
\begin{subfigure}{.49\textwidth}
        \centering
        Symmetric-50\%
        \includegraphics[width=\linewidth]{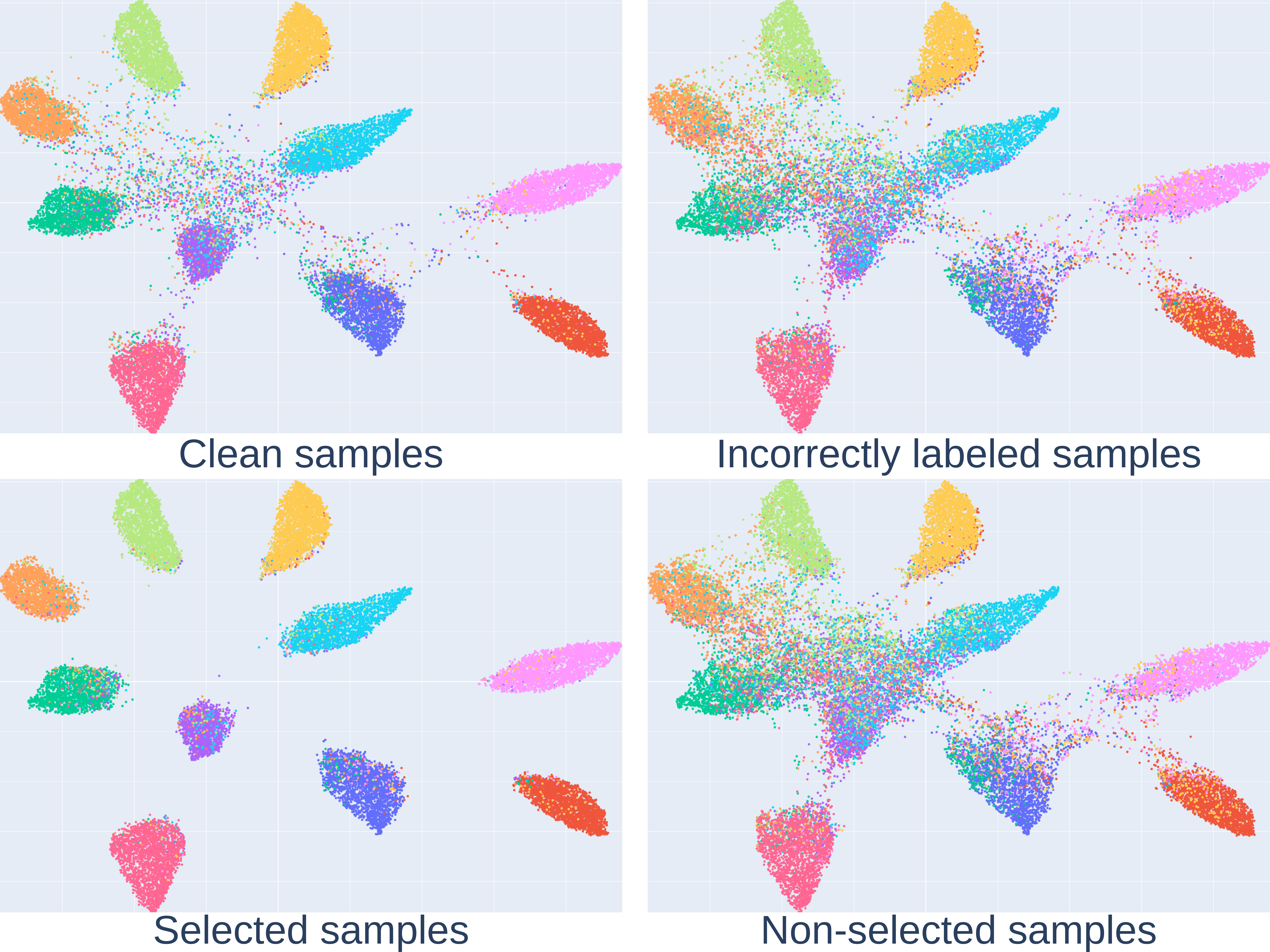}
\end{subfigure}\hfill
\par\bigskip
\begin{subfigure}{.49\textwidth}
        \centering
        Symmetric-80\%
        \includegraphics[width=\linewidth]{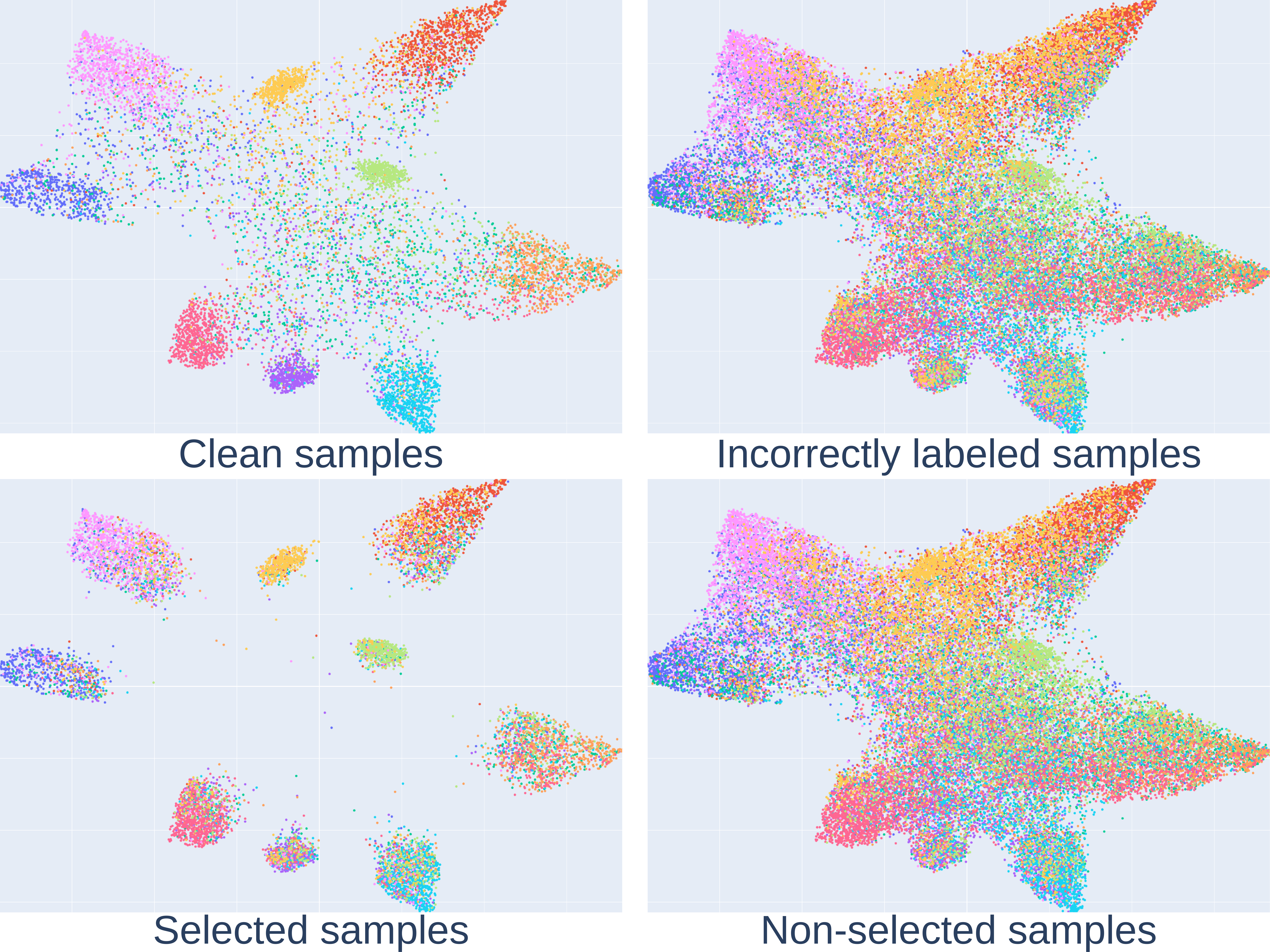}
\end{subfigure}\hfill
\begin{subfigure}{.49\textwidth}
        \centering
        Asymmetric-40\%
        \includegraphics[width=\linewidth]{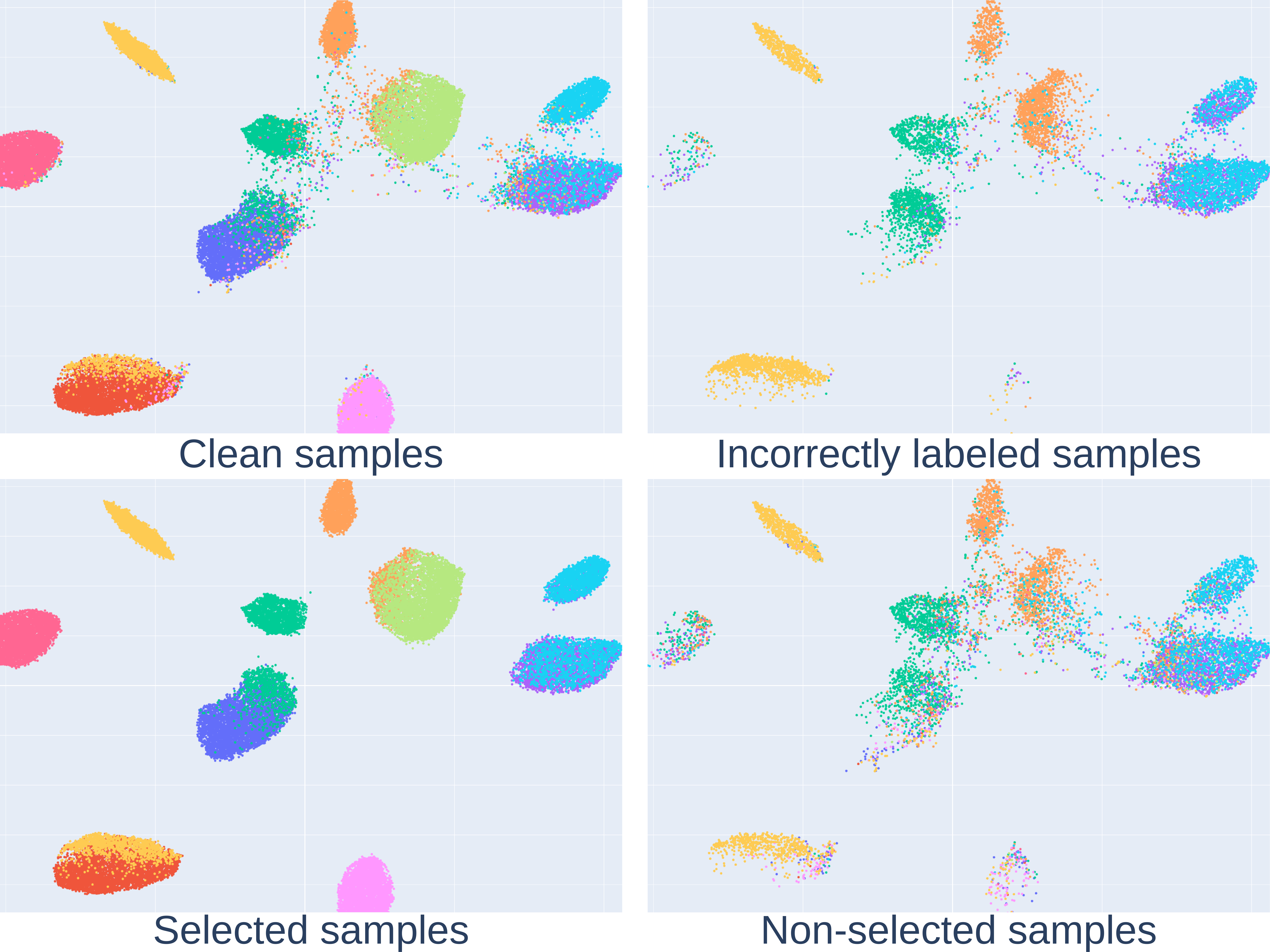}
\end{subfigure}\hfill
\caption{Visualization of the feature distribution of the noisy training samples on CIFAR-10.}
\label{fig:append_cifar10_samples}
\end{figure*}

\begin{figure*}[t]
\centering
\begin{subfigure}{.49\textwidth}
        \centering
        Symmetric-20\%
        \includegraphics[width=\linewidth]{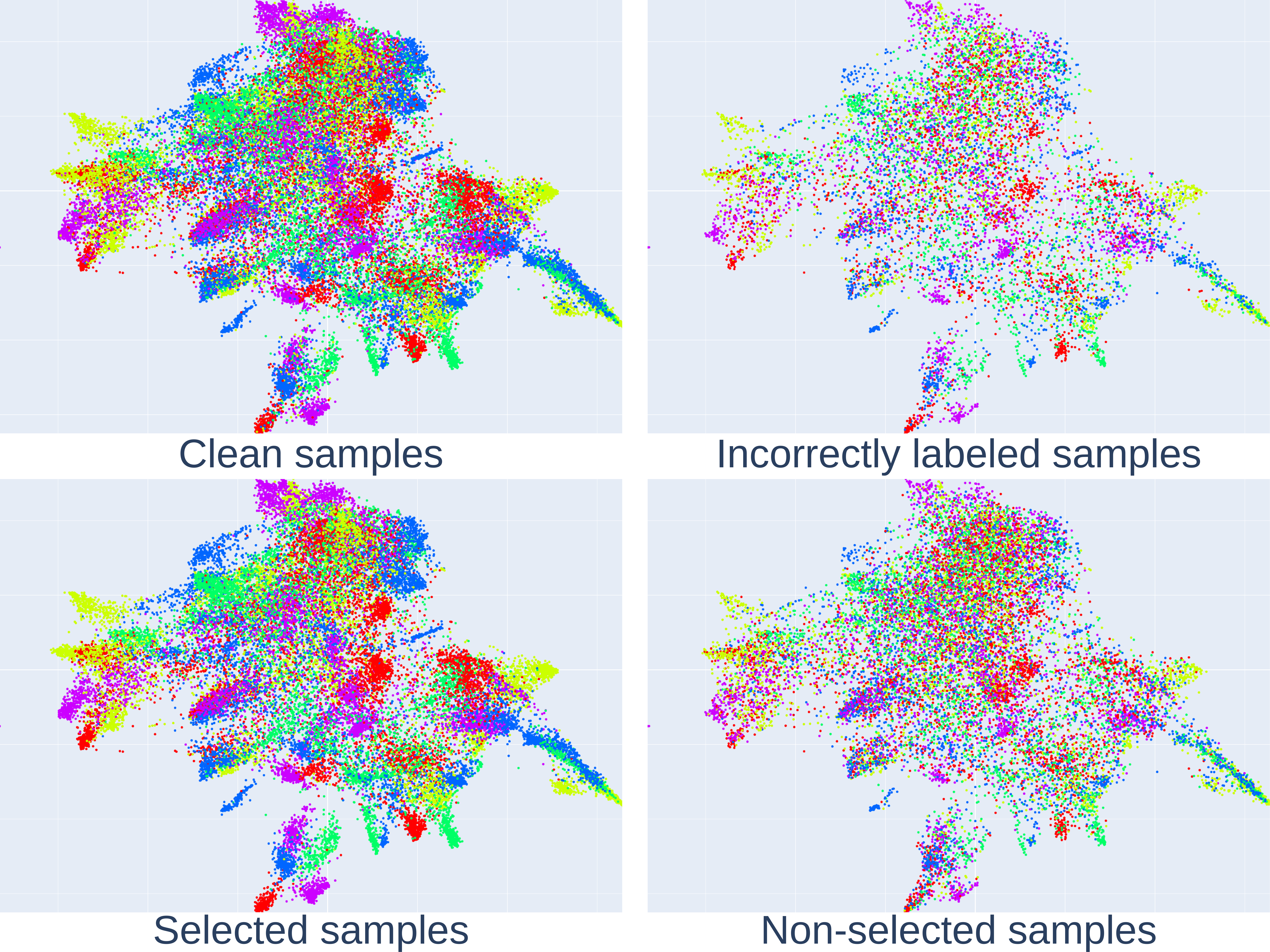}
\end{subfigure}\hfill
\begin{subfigure}{.49\textwidth}
        \centering
        Symmetric-50\%
        \includegraphics[width=\linewidth]{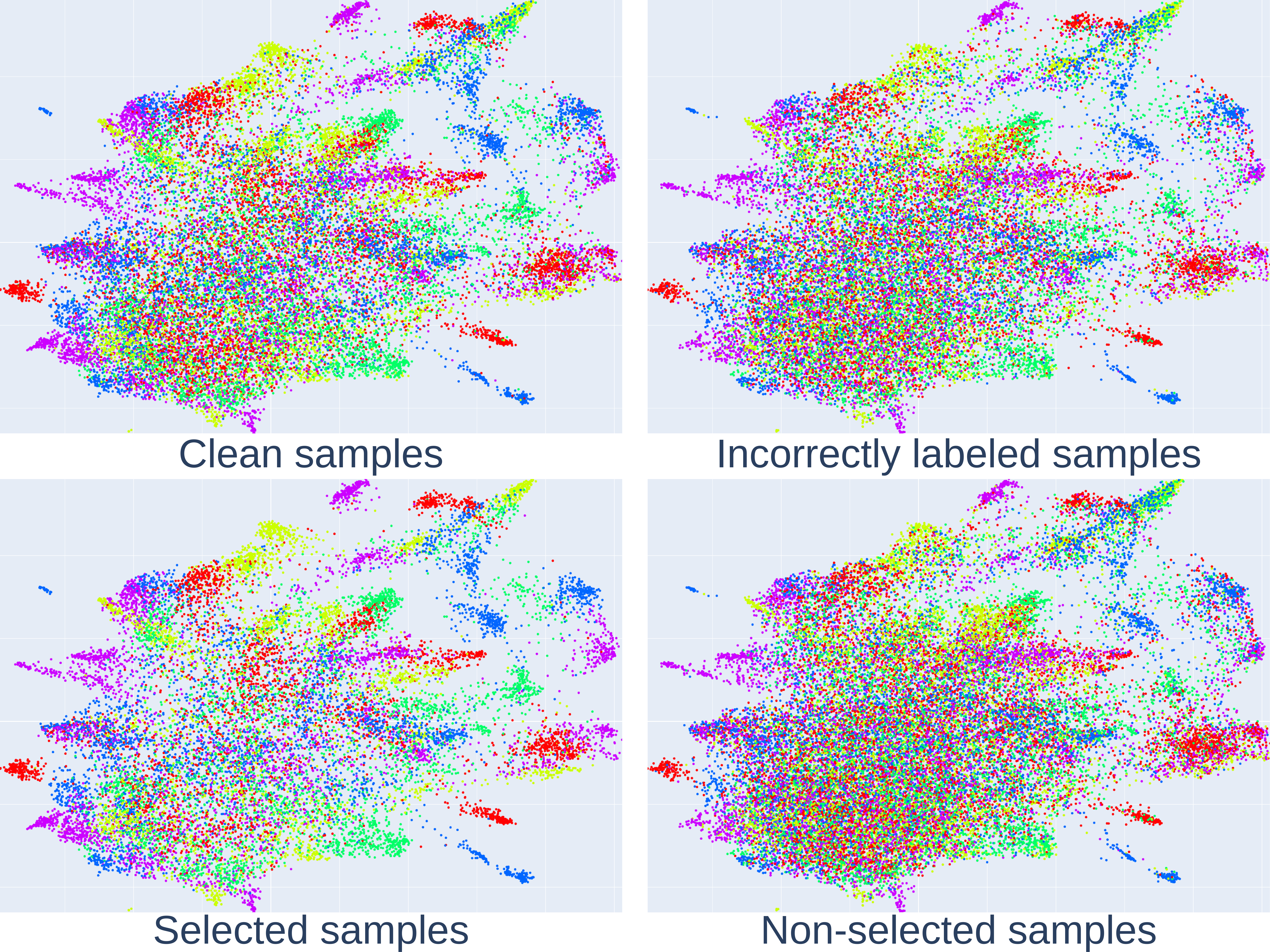}
\end{subfigure}\hfill
\par\bigskip
\begin{subfigure}{.49\textwidth}
        \centering
        Symmetric-80\%
        \includegraphics[width=\linewidth]{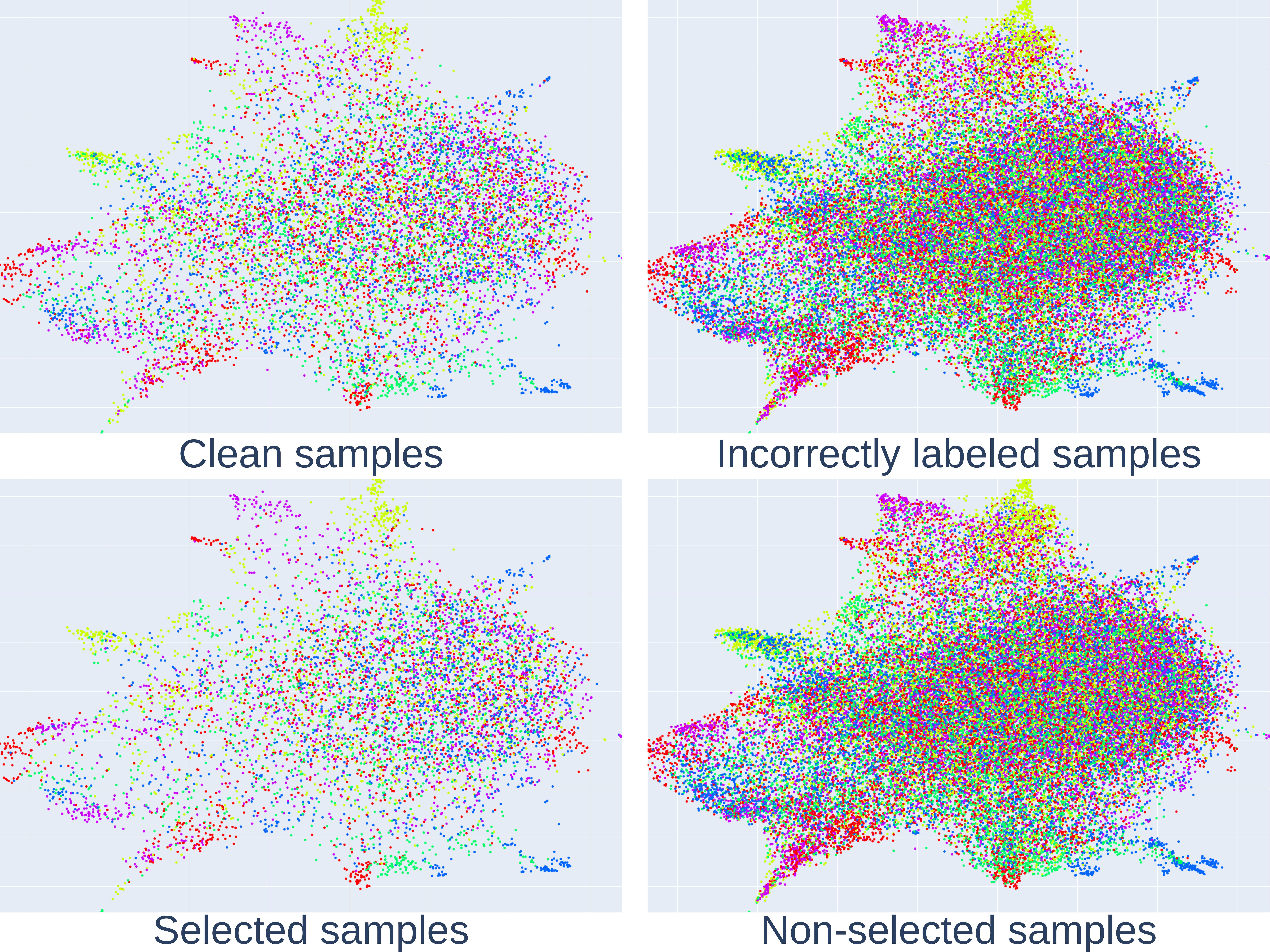}
\end{subfigure}\hfill
\begin{subfigure}{.49\textwidth}
        \centering
        Asymmetric-40\%
        \includegraphics[width=\linewidth]{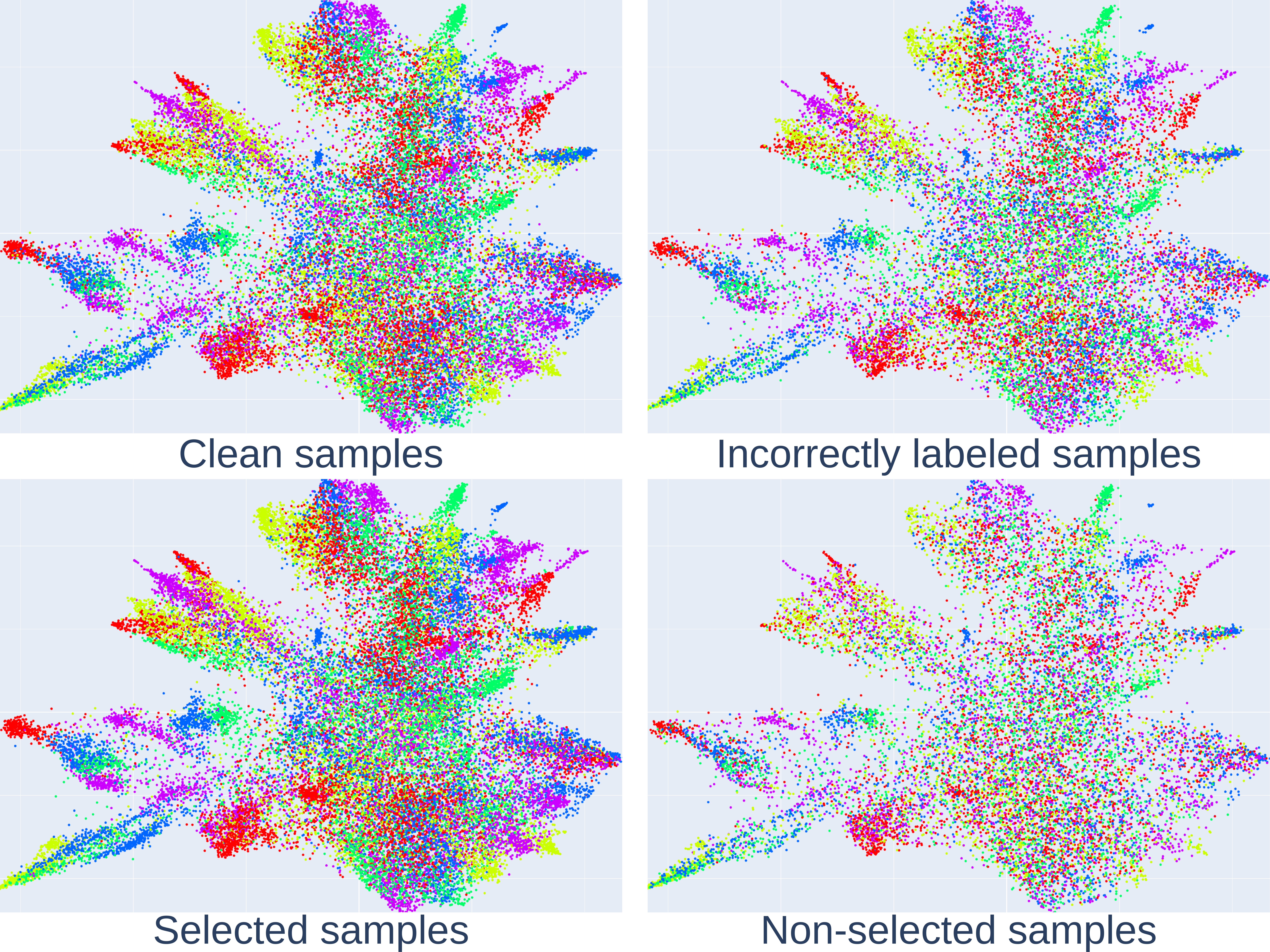}
\end{subfigure}\hfill
\caption{Visualization of the feature distribution of the noisy training samples on CIFAR-100.}
\label{fig:append_cifar100_samples}
\end{figure*}

Figs.~\ref{fig:append_mnist_samples},~\ref{fig:append_cifar10_samples}, and~\ref{fig:append_cifar100_samples} visualize the feature distributions of \Ours on MNIST, CIFAR-10, and CIFAR-100, respectively, by UMAP.
In the symmetric noise cases, incorrectly labeled samples tended to be scattered, and \Ours can ignore them successfully. In the asymmetric noise cases, \Ours selects most samples from the noiseless classes (e.g., ``pink'', ``violet-red'', ``red'', ``light-green'', and ``dark blue'' classes of CIFAR-10) and less samples from the noisy classes.

\section{Future work}
In future work, it would be interesting to design other effective noise-risk vector to estimate the noise-risk of samples more accurately. In addition, the idea of adaptive $k$-set selection could be applied to other learning tasks (e.g., ranking, regression, and detection) and datasets that contain noisy samples, such as samples from non-target classes and incomplete samples, rather than label noise.



\end{document}